\documentclass[10pt,twocolumn,letterpaper]{article}

\usepackage{iccv}              
\definecolor{iccvblue}{rgb}{0.21,0.49,0.74}
\usepackage[pagebackref,breaklinks,colorlinks,allcolors=iccvblue]{hyperref}
\def\paperID{7014} 
\def\confName{ICCV}
\def\confYear{2025}

\title{WaveMamba: Wavelet-Driven Mamba Fusion for RGB-Infrared Object Detection}

\author{
Haodong Zhu$^{1*}$ \quad
Wenhao Dong$^{1*}$ \quad
Linlin Yang$^{2\dagger}$ \quad
Hong Li$^1$ \quad
Yuguang Yang$^1$ \quad\\
Yangyang Ren$^1$ \quad
Qingcheng Zhu$^1$ \quad
Zichao Feng$^1$ \quad
Changbai Li$^1$ \quad
Shaohui Lin$^4$ \quad\\
Runqi Wang$^3$ \quad
Xiaoyan Luo$^{1\dagger}$ \quad 
Baochang Zhang$^{1}$ \quad\\
$^1$Beihang University, China \\
$^2$Communication University of China, China \\
$^3$Beijing Jiaotong University, China \\
$^4$East China Normal University, China \\
}

\begin{document}
\maketitle
\begin{abstract}
Leveraging the complementary characteristics of visible (RGB) and infrared (IR) imagery offers significant potential for improving object detection. In this paper, we propose WaveMamba, a cross-modality fusion method that efficiently integrates the unique and complementary frequency features of RGB and IR decomposed by Discrete Wavelet Transform (DWT). An improved detection head incorporating the Inverse Discrete Wavelet Transform (IDWT) is also proposed to reduce information loss and produce the final detection results. The core of our approach is the introduction of WaveMamba Fusion Block (WMFB), which facilitates comprehensive fusion across low-/high-frequency sub-bands. Within WMFB, the Low-frequency Mamba Fusion Block (LMFB), built upon the Mamba framework, first performs initial low-frequency feature fusion with channel swapping, followed by deep fusion with an advanced gated attention mechanism for enhanced integration. High-frequency features are enhanced using a strategy that applies an ``absolute maximum" fusion approach. These advancements lead to significant performance gains, with our method surpassing state-of-the-art approaches and achieving average mAP improvements of $4.5$\% on four benchmarks. 
\end{abstract}
\let\thefootnote\relax
\footnotetext{$^*$ Equal contribution.~~~\{HaodongZhu, ZB2315207\}@buaa.edu.cn}
\footnotetext{$^\dagger$ Corresponding author.}
\section{Introduction}
\label{sec:intro}

\begin{figure}[tbp]
    \includegraphics[width=0.5\textwidth]{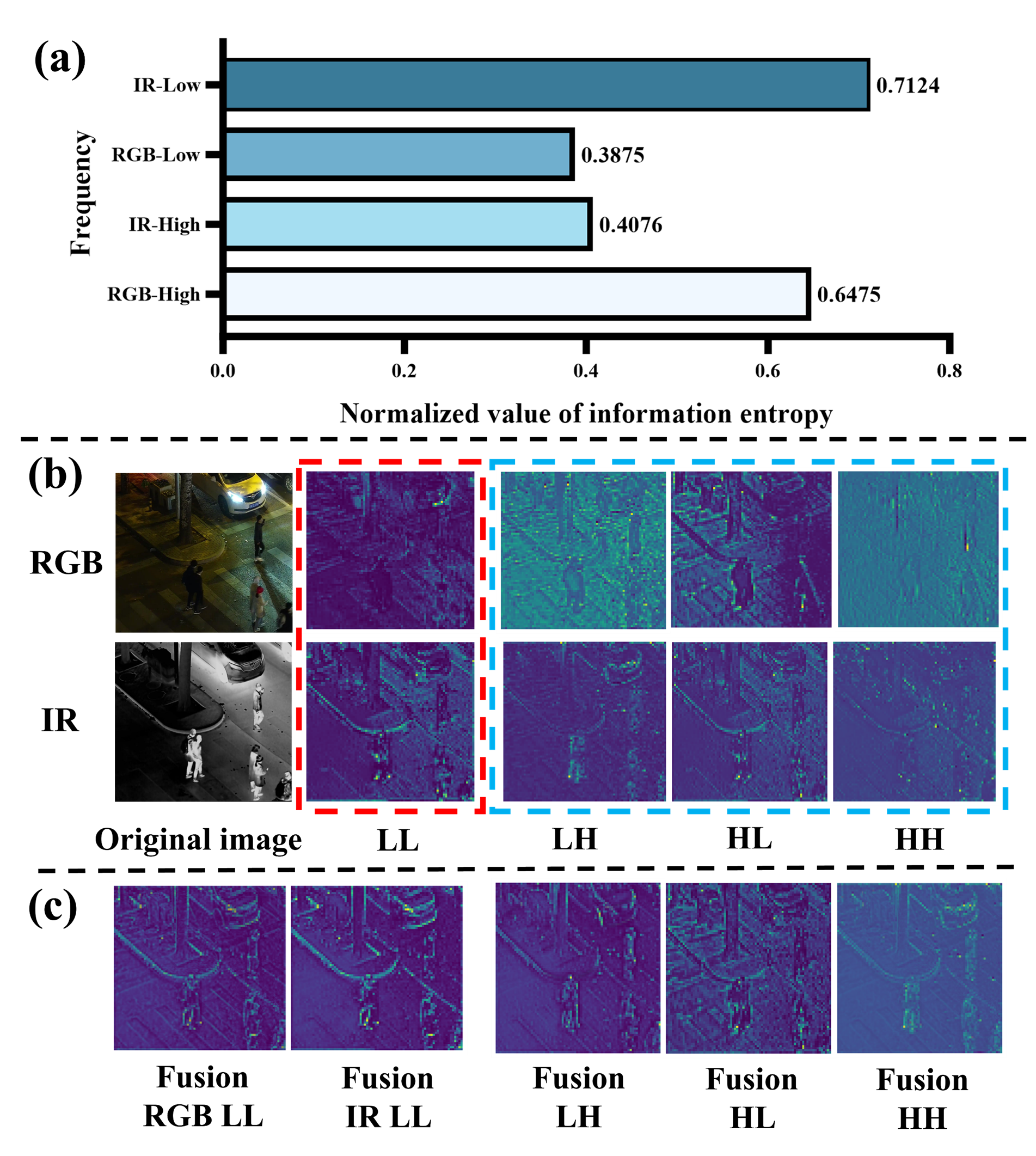}
    \caption{(a) shows the averaged normalized information entropy of low- and high-frequency components from RGB and IR images in $M^3$FD val dataset using DWT. 
    (b) depicts the RGB and IR features filtered by DWT in two directions, with low-frequency (L) and high-frequency (H) components obtained in each direction, resulting in four feature sub-bands: LL, LH, HL, and HH. Red and blue dashed boxes highlight the low-frequency and high-frequency components, respectively.
    The first row shows RGB and the second row shows IR components.
    (c) shows the result of fusion of each frequency.}
    
    \label{fig:1} 
    \vspace{-20pt}
\end{figure}

The fusion of multi-spectral features from visible (RGB) and infrared (IR) imagery has emerged as a pivotal technique in computer vision. This is especially applicable to scenarios in challenging environments where traditional methods alone may be insufficient~\cite{KimPR22, WangTLZL24, XieSCCJMH23}. RGB-based methods often encounter limitations in low-light or adverse weather conditions, as RGB images tend to lose detail and contrast. Conversely, IR images excel in these conditions by capturing thermal signatures that are independent of visible light, thus offering complementary information~\cite{QINGYUN2022108786,HUANG2024106156,ZHANG201920}. However, the integration of these distinct modalities (\ie, RGB and IR) into a cohesive framework poses significant challenges, including the effective fusion of heterogeneous data and maintaining computational efficiency.

Over the past decades, numerous approaches have been developed to integrate information from visible and infrared images for cross-modality object detection. These approaches can be typically classified into CNN-based~\cite{LiuZWM16, DBLP:journals/tamd/XieZYX23, rs14092020,s24196181}, Transformer-based~\cite{DBLP:journals/corr/abs-2111-00273, DBLP:journals/pr/ShenCLZFY24,10.1007/978-3-031-25072-9_41,10472947}, and Mamba-based~\cite{MambaST, zhou2024dmmdisparityguidedmultispectralmamba, ma2024s4fusionsaliencyawareselectivestate} methods. 
While CNN-based methods have demonstrated success, they often struggle to capture long-range dependencies, leading to a growing interest in Transformer-based techniques. Recently, the high computational complexity of Transformer models has further motivated the development of Mamba-based methods, offering a more efficient alternative. For instance, Gao et al.~\cite{MambaST} introduced MambaST, an efficient cross-spectral spatial-temporal fusion pipeline that enhances pedestrian detection by employing a multi-head hierarchical patching and aggregation structure to fuse visible and thermal imagery. 
Therefore, we utilize the Mamba framework~\cite{liu2024vmamba} as the foundation for our feature fusion due to its ability to balance efficiency and performance.

While existing cross-modality object detection methods have shown promising performance, 
research on effectively using inherent features between modalities for complementary advantages to enhance performance remains insufficient.
We endeavor to explore answers from the frequency domain.
Specifically, as shown in Fig.~\ref{fig:1}(a), applying Discrete Wavelet Transform (DWT) to $M^3$FD validation set reveals that IR images exhibit higher normalized information entropy in the low-frequency sub-band, while RGB images exhibit greater entropy in the high-frequency sub-bands. The value of information entropy reflects the amount of information carried by the image. 
Generally speaking, the higher the information entropy, the richer the content of image is~\cite{wang2022imagerestorationqualityassessment}. 
This statistical observation suggests that RGB images are more adept at capturing high-frequency details, in contrast to IR images, which predominantly encapsulate low-frequency signals. 
Furthermore, Fig.~\ref{fig:1}(b) illustrates the DWT outputs of P2-level RGB and IR features derived from the YOLOv8 backbone trained on each respective modality. The low- and high-frequency components are delineated by the red and blue dashed box, respectively. 
In the low-frequency region, IR features manifest with clearer shapes, while RGB features present more pronounced edges and contours in the high-frequency region, affirming their distinct frequency characteristics—an observation that aligns with prior studies ~\cite{DBLP:journals/tim/WuHYZRLXZB23}. 
Building on these insights, \emph{we aim to exploit the unique and complementary frequency characteristics of IR and RGB modalities to achieve enhanced feature fusion,  thereby improving object detection performance.}

\begin{figure*}[tbp]
    \centering
    \includegraphics[width=0.90\linewidth]{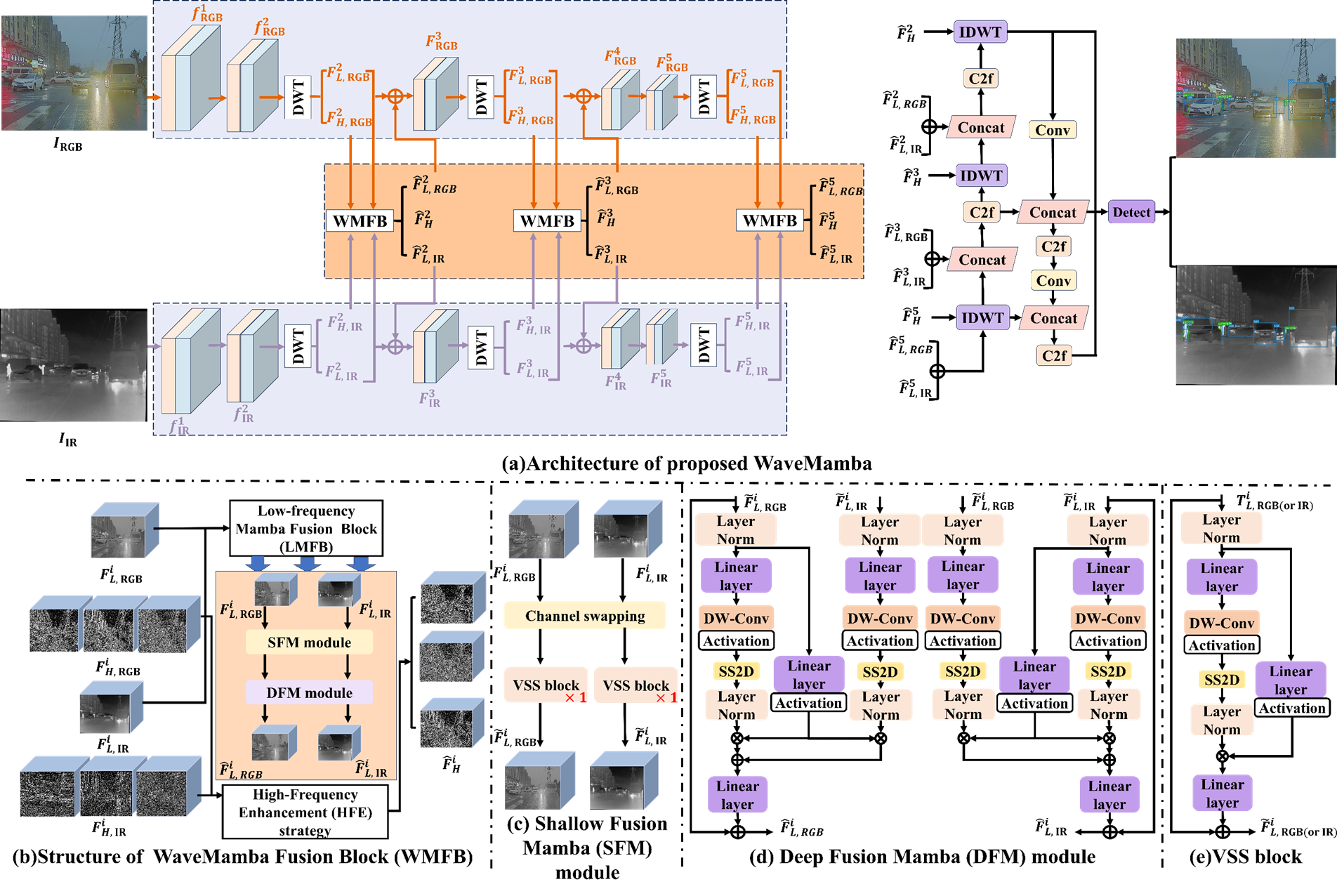}
    \vspace{-1em}
    \caption{Illustrative diagram of the complete network and sub-modules. (a) shows the overall architecture of proposed WaveMamba network, which mainly includes the dual-stream feature extraction backbone improved by adding DWT, three WaveMamba Fusion Blocks and the improved YOLOv8 head augmented with IDWT.  
    (b) shows the detailed structure of the WMFB, which includes the LMFB and the HFE strategy. (c), (d) and (e) show the detailed structure of the SFM module, DFM module and VSS block, respectively.
    }
    \label{fig:main}
    \vspace{-10pt}
\end{figure*}

In this paper, we propose WaveMamba, as illustrated in Fig.~\ref{fig:main}, which leverages the DWT to map features into the wavelet domain, exploiting the complementary frequency characteristics of RGB and IR for efficient feature fusion. An improved detection head with Inverse DWT (IDWT) is then used to generate the final detection results, leading to improved detection accuracy. 
The main innovation of WaveMamba lies in WaveMamba Fusion Block (WMFB), which comprises two key components: the Low-frequency Mamba Fusion Block (LMFB) and the High-Frequency Enhancement (HFE) strategy. 
LMFB utilizes the exceptional low-frequency modeling capability of Mamba ~\cite{ma2024tinyvim} to integrate low-frequency information. It comprises two modules: the Shallow Fusion Mamba (SFM) module, which facilitates lightweight cross-modality interaction by swapping portions of RGB and IR channels,
and the Deep Fusion Mamba (DFM) module, which further refines the fusion process using a gated attention mechanism to filter out redundant information and improve the quality of the low-frequency feature fusion. Meanwhile, the HFE strategy enhances the selection and retention of critical details within high-frequency features by adopting an ``absolute maximum'' fusion approach. 
This combination of innovations significantly improves the fusion of features from both modalities, as shown in Fig.~\ref{fig:1}(c). 
The low-frequency features of RGB fusion are enriched, while the IR fusion features effectively suppress redundant noise in the originally low-frequency characteristics of IR, such as road shadows. 
Meanwhile, the fused high-frequency features become more detailed and pronounced.
In summary, this paper's main contributions are:

\begin{itemize}
    \item We propose WaveMamba, which leverages the intrinsic advantageous frequency characteristics of different modalities to fuse features, facilitating sufficient feature interaction and enhancing the performance of RGB-IR object detection.
    
    \item WaveMamba Fusion Block efficiently integrates distinct frequency information, including a Low-frequency Mamba Fusion Block that combines low-frequency features and a High-Frequency Enhancement strategy to capture details without adding computational complexity. An improved YOLOv8 detection head utilizes IDWT for final detection results.

    \item Extensive experiments on four public datasets demonstrate that our approach
    achieves state-of-the-art performance in cross-modality object detection tasks, with an average $m$AP exceeding that of the second-best method by \(4.5\%\). 

\end{itemize}

\section{Related Works}
\label{sec:formatting}
\textbf{Multi-modality Object Detection.} Various multi-modality object detection methods have been proposed based on the complementary information from RGB-IR images. So far, these methods can be categorized into two main directions: pixel-based and feature-based. Pixel-based methods such as CDDFuse~\cite{Zhao_2023_CVPR} and SuperFusion~\cite{9970457}, involve fusing two images at the pixel level to obtain a single fused image, upon which object detection is then performed. Feature-based methods fuse the output from a certain stage of the two-stream detector, such as the early or middle stage features extracted by the backbone~\cite{rs14092020,10209020,9739079,10114594,10382506}
or the results of the detector (\ie, late fusion)~\cite{LI2019161,10.1007/978-3-031-20077-9_9,10225383,10478590}. 
However, the fusion modules designed in these methods are predominantly based on CNN or Transformer architectures.
CNN's are constrained by their local receptive fields, while Transformer's face inefficiencies in model detection due to their complexity.  
Recently, the DMM framework, as described in \cite{zhou2024dmmdisparityguidedmultispectralmamba}, incorporates fusion modules designed to optimize and merge features from two modalities using Mamba, leading to successful outcomes in oriented object detection.
Despite the excellent performance achieved by these methods, further enhancement in detection performance can be realized by utilizing distinct characteristics in the frequency domain between modalities.\\
\textbf{Wavelet Transform in Computer Vision.} As wavelet transform can achieve lossless up- and down-sampling and obtain frequency domain information of images, it has been widely applied to various computer vision tasks~\cite{10.1007/978-3-031-19806-9_19,TIAN2023109050,HUANG2022104737,Zou_2021_ICCV}, resulting in performance improvements. WTConv~\cite{finder2024waveletconvolutionslargereceptive} achieves a significant increase in the convolutional receptive field by integrating wavelet and inverse wavelet transforms into the convolution process. WF-Diff~\cite{10655245} attains cutting-edge performance in underwater image restoration by using wavelet transform to obtain the high- and low-frequency components of the image after preliminarily enhanced, and then sending them into two separate diffusion branches to adjust the high- and low-frequency information. 
DCFNet~\cite{DBLP:journals/sensors/WuWWWG24} realizes an improvement in the quality of fused images for the task of RGB and infrared image fusion. It incorporates wavelet transform and inverse wavelet transform as pre-processing steps for the Encoder and Decoder, respectively. 
However, these methods fail to distinguish different frequency domain characteristics of low- and high-frequency components, adopting either identical processing methods for both or without processing the high-frequency components altogether. Consequently, the extraction of frequency domain information is inadequate.\\
\indent Unlike the aforementioned methods, we utilize wavelet transform and its inverse transform in cross-modality object detection, designing distinct processing modules for high- and low-frequency components, resulting in significant improvements in detection efficiency and performance.

\section{Methods}
\subsection{Preliminaries}
\label{sec:Preliminaries}
\noindent\textbf{Discrete Wavelet Transform.}  
Haar wavelet transform is a standard DWT due to its simplicity and efficiency, and we set it as the default. 
For an image $I$
, the one-level transform applies depth-wise convolution using low-pass \( L = \frac{1}{\sqrt{2}}[1, 1] \) and high-pass \( H = \frac{1}{\sqrt{2}}[1, -1] \) filters, followed by a down-sampling step that reduces the resolution by a factor of 2. Extending to two dimensions involves four filters, and we have sub-bands for $I$ as below: 
\begin{equation}  
\label{WT2D}  
\begin{aligned}  
F_{L}, F_{H} = \text{DWT}(I), 
\end{aligned}  
\end{equation}
where $\text{DWT}(\cdot)$ denotes the Haar wavelet transform, sub-bands $F_{L} = \{F_{LL}\}$ captures low-frequency information and $F_{H} = \{F_{LH},F_{HL},F_{HH}\}$ captures high-frequency details and noise.

\noindent\textbf{2D Selective Scan.}
In linear time-invariant systems, state space models (SSMs) process 1D input signals through state variables and genetate outputs using ODEs.
Mamba~\cite{gu2023mamba} enhances this modeling approach, but its direct application to vision tasks is limited by the mismatch between 2D visual data and 1D language sequences. 
To address this, the 2D Selective Scan (SS2D) mechanism introduced in~\cite{liu2024vmamba} expands image patches in four directions, creating independent sequences. Each feature sequence is processed using the Selective Scan Space State Sequential Model (S6)~\cite{gu2023mamba}, and the sequences are then aggregated to reconstruct the 2D feature map.

\subsection{WaveMamba}

\label{sec:WaveMambafusion}

The architecture of our model, as depicted in Fig.~\ref{fig:main}, consists of a dual-stream feature extraction backbone augmented with the DWT, three WaveMamba Fusion Blocks, and an improved YOLOv8 detection head that incorporates the IDWT.

Given the input RGB images $I_{\text{RGB}}$ and infrared images $I_{\text{IR}}$,  our model utilizes the first two layers of the dual-stream backbone to extract local features \( f^i_{\text{RGB}} \) and \( f^i_{\text{IR}} \) where the superscript $i \in \{1,2\}$ indicates the layer. \( f^2_{\text{RGB}} \) and \( f^2_{\text{IR}} \) undergo the Haar wavelet transformation and result in corresponding features in the wavelet domain:

\begin{equation}
\begin{aligned}
    &F^{2}_{L, \text{RGB}},F^{2}_{H, \text{RGB}} = \text{DWT}(f^2_{\text{RGB}}),\\
    &F^{2}_{L, \text{IR}},F^{2}_{H, \text{IR}} = \text{DWT}(f^2_{\text{IR}}),
\end{aligned}
\label{eq4}
\end{equation}
\\
where $F^{2}_{L, \text{RGB}}$ and $F^{2}_{L, \text{IR}}$ denote the low-frequency sub-bands, while $F^{2}_{H, \text{RGB}}$ and $F^{2}_{H, \text{IR}}$ represent the high-frequency sub-bands of the RGB and infrared modalities, respectively.
With these frequency features combined with our proposed WMFB, we obtain fused features $F^3_{\text{RGB}}$ and $F^3_{\text{IR}}$.  Like~\cite{finder2024waveletconvolutionslargereceptive}, multi-level wavelet transformations are used in our framework to achieve more detailed multi-scale features. Specifically, DWT and WMFB are applied to the fused features at the $3^{rd}$ and $5^{th}$ layers.
Finally, the fused low- and high-frequency sub-bands (\ie, all the output of WMFB) are fed into the improved YOLOv8 detection head, where IDWT is used for feature up-sampling, resulting in the final detection outputs.
In the following sections, we introduce WMFB in subsec.~\ref{subsec:wmfb} and the improved YOLOv8 head in subsec.~\ref{subsec:yolo}.

\subsection{WaveMamba Fusion Block}
\label{subsec:wmfb}
As shown in Fig.~\ref{fig:1}, we find that RGB images typically possess more high-frequency details, while IR images primarily capture low-frequency information. 
To effectively leverage the complementary frequency characteristics of both RGB and IR for feature fusion, we develop the WMFB. 
Specifically, WMFB consists of the Low-frequency Mamba Fusion Block and the High-Frequency Enhancement strategy for low- and high-frequency fusion, respectively.

\subsubsection{Low-frequency Mamba Fusion Block}
LMFB processes low-frequency features $F^{i}_{L, \text{RGB}}$, $F^{i}_{L, \text{IR}}$ in $i^{th}$ layer to integrate low-frequency information from shallow to deep. 
In this case, LMFB designs the shallow fusion mamba module and deep fusion mamba module.

\noindent\textbf{SFM Module.} 
This module is designed to enhance the interaction of low-frequency features across modalities for shallow feature fusion. It utilizes a channel swapping operation~\cite{Panmamba} in conjunction with VSS blocks~\cite{liu2024vmamba}.
We first integrate information from distinct channels to construct cross-modality feature correlations, thereby enriching the diversity of channel features and improving fusion performance. We begin with the channel swapping to derive new local features for low-frequency RGB and IR modalities, respectively. This can be formulated as follows:
\begin{equation}
\begin{aligned}
    T^{i}_{L,\text{RGB}},T^{i}_{L,\text{IR}} =\text{CS}(F^{i}_{L, \text{RGB}},F^{i}_{L, \text{IR}}), 
\end{aligned}
\label{eq5}
\end{equation}
where \( \text{CS}(\cdot) \) denotes the channel swapping operation. This operation is implemented through channel splitting and concatenation to enhance the diversity of channel features. 
To further enhance the fused features, two VSS blocks are applied to  \( T^{i}_{L,\text{RGB}} \) and \( T^{i}_{L,\text{IR}} \) separately as below:
\begin{equation}
\begin{aligned}
    &\tilde{F}^{i}_{L, \text{RGB}}=\text{VSS}(T^{i}_{L,\text{RGB}}), \quad \tilde{F}^{i}_{L, \text{IR}}=\text{VSS}(T^{i}_{L,\text{IR}}),
\end{aligned}
\label{eq6}
\end{equation}
where \( \text{VSS}(\cdot) \) refers to the VSS block, as shown in Fig.~\ref{fig:main}(e). The outputs \( \tilde{F}^{i}_{L, \text{RGB}} \) and \( \tilde{F}^{i}_{L, \text{IR}} \) represent the shallow fused features from the RGB and IR modalities, respectively.

\noindent\textbf{DFM Module.}
To further facilitate the interaction and deep fusion of low-frequency features between modalities, we introduce DFM based on the output of SFM as shown in Fig.~\ref{fig:main}(d).
In DFM, the SFM outputs $(\tilde{F}^{i}_{L, \text{RGB}}, \tilde{F}^{i}_{L, \text{IR}})$ alternate between primary and auxiliary roles. The primary modality is linearly embedded and split into two streams: one passes through a 3×3 depth-wise convolution, SiLU activation, the core SS2D module, and layer normalization; the other undergoes only SiLU activation. The auxiliary modality follows the primary stream's first pathway.
Given the complexity and diversity of features present across different modalities, it is crucial to selectively emphasize unique and complementary characteristics while minimizing redundancy.
On this basis, we introduce a gated attention mechanism to effectively facilitate the learning of complementary features while suppressing redundant features. Specifically, this mechanism uses the output of the second information flow of the primary modality to regulate the outputs of both the first information flow of the primary modality and the auxiliary modality. By summing these regulated results, we obtain the final output of the DFM module for $i^{th}$ layer, formalized as follows:  
\begin{equation}
\begin{aligned}
    &\hat{F}^{i}_{L, \text{RGB}}, \hat{F}^{i}_{L, \text{IR}} = \text{DFM}(\tilde{F}^{i}_{L, \text{RGB}}, \tilde{F}^{i}_{L, \text{IR}}).
\end{aligned}
\label{eq11}
\end{equation}

\subsubsection{High-Frequency Enhancement Strategy}
High-frequency sub-bands capture fine details in the original features, where larger absolute values indicate more significant information.
To select and retain critical details, we employ a HFE strategy, which efficiently integrates high-frequency information by choosing pixels with larger absolute values.  
The HFE strategy for $i^{th}$ layer high-frequency sub-bands is defined as follows:  
\begin{equation}  
\begin{aligned}  
    F^{i}_{H} &= \text{Mask}(|F^{i}_{H, \text{RGB}}|-|F^{i}_{H, \text{IR}}|) \odot F^{i}_{H, \text{RGB}} \\
               &\quad +\text{Mask}(|F^{i}_{H, \text{IR}}|-|F^{i}_{H, \text{RGB}}|) \odot F^{i}_{H, \text{IR}},  
\end{aligned}  
\label{eq13}  
\end{equation}  
where  \( \odot \) is element-wise product operation. \( \text{Mask}(\cdot) \) generates an element-wise binary mask, where element is set to 1 if its corresponding input is greater than 0; otherwise, it is set to 0.
Note that
$F^i_{H} = \{F^i_{LH},F^i_{HL},F^i_{HH}\}$ based on Eq.~\ref{WT2D}. In this case, we process $F^i_{LH},F^i_{HL},F^i_{HH}$ independently and construct \( F^{i}_{H} \). Lastly,
\( F^{i}_{H} \) is the resulting merged high-frequency subbands.

\subsection{Improved YOLOv8 head}
\label{subsec:yolo}
Since our network architecture utilizes multilevel wavelet transformations to obtain multi-scale features, we intuitively design a corresponding YOLOv8 detection head structure using IDWT.
Compared to traditional upsampling, IDWT reduces information loss and significantly improves the accuracy of detection results.  

As shown in Fig.~\ref{fig:main}(a), initially, we aggregate the fused low-frequency features from two modalities to get the final low-frequency feature. Subsequently, IDWT is utilized on this low-frequency feature alongside the fused high-frequency features. Aside from this modification, the rest in the detection head remain consistent with those in the original YOLOv8. Through these operations, we derive the ultimate detection results.

\section{Experiments}
\begin{table*}
  \centering
  \small
  \scalebox{0.75}{\begin{tabular}{cc|cc|cccccc}
    \toprule
    Methods & Backbone& $mAP_{50}$ & $mAP$ & Car & Bus &Lamp &People&Truck&Motorcycle\\
    \midrule
    (LOPET'24) LG+MDA~\cite{10.1117/12.3040116}&ResNet50&66.4&40.2&83.0&84.4&37.7&62.5&75.5&54.6\\
    (ECCV'24) DAMSDet~\cite{guo2024damsdet}&ResNet50&80.2&52.9&-&-&-&-&-&-\\
    Ours & ResNet50&\textcolor{blue}{90.9}&\textcolor{blue}{62.3}&\textcolor{green}{95.6}&\textcolor{red}{96.4}&\textcolor{blue}{92.8}&\textcolor{blue}{91.6}&84.8&84.4\\
    \midrule
    (CVPR'22) RFNet~\cite{9878923}&YOLOv5&79.4&53.2&91.1&78.2&85.0&79.4&69.9&72.8\\
    (CVPR'22) TarDAL~\cite{9879642}&YOLOv5&80.5&54.1&94.8&81.3&87.1&81.5&68.7&69.3\\
    (IJCAI'20) DIDFuse~\cite{Zhao_2020}&YOLOv5&77.2&51.4&91.0&78.3&83.2&78.5&66.3&66.3\\
    (IJCV'21) SDNet~\cite{Zhang2021SDNetAV}&YOLOv5&79.3&52.9&91.1&78.2&85.0&79.4&69.0&72.8\\
    (MM'22) DetFusion~\cite{10.1145/3503161.3547902}&YOLOv5&80.8&53.8&92.5&83.0&87.8&80.8&71.4&69.4\\
    (CVPR'23) CDDFuse~\cite{Zhao_2023_CVPR}&YOLOv5&80.5&53.6&91.6&82.4&86.0&81.1&71.2&71.1\\
    (MM'23) IGNet~\cite{10.1145/3581783.3612135}&YOLOv5&81.5&54.5&92.8&82.4&86.9&81.6&72.1&73.0\\
    (TCSVT'24) MMFN~\cite{10666754}&YOLOv5&86.2&57.4&93.2&92.1&87.6&83.0&87.4&73.7\\
    (CVPR'24) EMMA~\cite{Zhao_2024_CVPR}&YOLOv5&82.9&55.4&93.5&83.2&87.7&82.0&73.5&77.7\\
    (TIM'24) KCDNet~\cite{10632179}&YOLOv5&83.2&56.3&90.9&88.4&80.3&83.3&72.1&84.1\\
    (TJRS'24) CRSIOD~\cite{10440361}&YOLOv5&84.0&57.2&92.2&93.3&80.5&78.4&85.5&73.9\\
    Ours & YOLOv5&\textcolor{green}{91.5}&\textcolor{green}{63.8}&\textcolor{blue}{95.4}&95.4&\textcolor{green}{93.2}&\textcolor{green}{91.8}&\textcolor{blue}{87.8}&\textcolor{blue}{85.7}\\
    \midrule
    YOLOv8l-IR~\cite{10533619}&YOLOv8&79.5&53.1&90.0&90.9&63.0&82.9&85.9&64.6\\
    YOLOv8l-RGB~\cite{10533619}&YOLOv8&80.9&52.5&91.2&92.9&75.3&70.6&86.0&69.6\\
    \midrule
    (JAS'22) SuperFusion~\cite{9970457}&YOLOv7&83.5&56.0&91.0&93.2&70.0&83.7&85.8&77.4\\
    (2024) TSJNet~\cite{jie2024tsjnetmultimodalitytargetsemantic}&YOLOv7&86.0&58.9&91.8&\textcolor{green}{95.7}&70.4&81.8&\textcolor{red}{89.3}&\textcolor{red}{86.8}\\
    (YAC'24) RI-YOLO~\cite{10598725}&YOLOv8&83.6&56.6&91.6&88.5&75.7&84.1&\textcolor{blue}{87.8}&73.5\\
    (Sensors'24) MRD-YOLO~\cite{s24103222}&YOLOv8&86.6&59.3&-&-&-&-&-&-\\
    Ours & YOLOv8&\textcolor{red}{92.1}&\textcolor{red}{64.4}&\textcolor{red}{95.9}&\textcolor{blue}{95.6}&\textcolor{red}{93.8}&\textcolor{red}{92.6}&\textcolor{green}{88.6}&\textcolor{green}{86.1}\\
    \bottomrule
  \end{tabular}}
  \caption{Comparison results with SOTA methods on $M^3$FD dataset. The best results are highlighted in \textcolor{red}{red}. The second and third best results are
highlighted in \textcolor{green}{green} and \textcolor{blue}{blue}, respectively.}
  \label{tab:example3}
\end{table*}
\begin{table*}
  \centering
  \small
  \scalebox{0.75}{\begin{tabular}{cc|cc|ccccc}
    \toprule
    Methods & Backbone& $mAP_{50}$ & $mAP$ & Car & Bus &Truck &Freight-car&Van \\
    \midrule
    (ECCV'22) TSFADet~\cite{10.1007/978-3-031-20077-9_30}&ResNet50&73.1&44.1&89.9&89.8&67.9&63.7&54.0\\
    (TGRS'24) $C^2$Former-$S^2$ANet~\cite{10472947}&ResNet50&74.2&47.3&90.2&89.8&68.3&64.4&58.5\\
    (GRSL'24) GLFNet~\cite{10476333}&ResNet50&71.4&54.8&90.3&88.0&72.7&53.6&52.6\\
    (JST'24) FFODNet~\cite{10461034}&ResNet50&76.3&56.9&-&-&-&-&-\\
    (2024) DMM~\cite{zhou2024dmmdisparityguidedmultispectralmamba}&ResNet50&77.2&55.8&90.4&88.7&77.8&63.0&\textcolor{red}{66.0}\\
    (TGRS'24) LF-MDet~\cite{10643097}&ResNet50&71.8&51.3&-&-&-&-&-\\
    Ours & ResNet50&\textcolor{blue}{79.3}&\textcolor{blue}{59.9}&94.6&90.2&\textcolor{blue}{79.6}&\textcolor{blue}{68.0}&64.1\\
    \midrule
    (YAC'24) CAFN-IA~\cite{10598791}&YOLOv5&69.3&56.1&89.1&90.8&62.0&57.3&47.1\\
    (MTA'23) SLBAF-Net~\cite{10.1007/s11042-023-15333-w}&YOLOv5&76.8&49.5&90.2&89.9&76.0&\textcolor{green}{68.2}&59.9\\
    (CVPR'24) CSOM-ODAF~\cite{10655285}&YOLOv5&77.1&50.1&90.1&89.8&75.6&\textcolor{green}{68.2}&61.8\\
    (RSL'24) multimodal DINO~\cite{article}&YOLOv5& 72.5&50.3&89.5&88.8&75.4&54.3&54.3\\
    Ours & YOLOv5&\textcolor{green}{79.5}&\textcolor{green}{60.2}&94.8&90.1&\textcolor{green}{80.1}&\textcolor{green}{68.2}&\textcolor{blue}{64.3}\\
    \midrule
    YOLOv8l-IR~\cite{10533619}&YOLOv8&71.9&49.1&93.4&91.9&69.3&53.7&51.1\\
    YOLOv8l-RGB~\cite{10533619}&YOLOv8&70.2&48.6&92.5&91.7&68.8&47.8&50.2\\
    \midrule
    (Sensors'23) Dual-YOLO~\cite{s23062934}& YOLOv7&71.9&55.2&\textcolor{green}{95.9}&91.6&69.7&55.9&46.6\\
    (AEORS'24) YOLOFIV~\cite{10643643}&YOLOv8&64.7&53.1&\textcolor{green}{95.9}&91.6&64.2&34.6&37.3\\
    (Sensors'24) IV-YOLO~\cite{s24196181}&YOLOv8&74.6&56.8&\textcolor{red}{97.2}&\textcolor{red}{94.3}&65.4&63.1&53.0\\
    (RS'24) DAAB-FFPN~\cite{rs16203904}&YOLOv8&75.2&56.3&-&-&-&-&-\\
    (TGRS'24) CRSIOD~\cite{10440361}&YOLOv8&73.2&50.8&95.6&\textcolor{blue}{92.2}&71.7&50.5&55.8\\
    (IJAEOG’24) CMA~\cite{JIANG2024103918}&YOLOv8&76.8&50.4&\textcolor{blue}{95.8}&\textcolor{green}{93.1}&75.9&59.8&59.4\\
    Ours & YOLOv8&\textcolor{red}{79.8}&\textcolor{red}{60.5}&95.0&90.6&\textcolor{red}{80.4}&\textcolor{red}{68.5}&\textcolor{green}{64.5}\\
    \bottomrule
  \end{tabular}}
  \caption{Comparison results with SOTA methods on DroneVehicle dataset. The best results are highlighted in \textcolor{red}{red}. The second and third best results are
highlighted in \textcolor{green}{green} and \textcolor{blue}{blue}, respectively.}
  \label{tab:example4}
  \vspace{-10pt}
\end{table*}

\begin{table}
  \centering
  \small
  \scalebox{0.75}{
  \begin{tabular}{cc|cc}
    \toprule
    Methods&Backbone&$mAP_{50}$ & $mAP$\\
    \midrule
    (CVPRW'23) CSAA~\cite{10209020}&ResNet50&94.3&54.2\\
    (2024) RSDet~\cite{zhao2024removalselectioncoarsetofinefusion}&ResNet50&95.8&61.3\\
    (TIV'24) MMSANet~\cite{10648743}&ResNet50&96.4&62.5\\
    (CVPR'24) Text-IF~\cite{Yi_2024_CVPR}&Transformer&94.1&60.2\\
    Ours & ResNet50&\textcolor{blue}{97.3}&\textcolor{blue}{65.0}\\
    \midrule
    (IF’23) DIVFusion~\cite{TANG2023477}&YOLOv5&89.8&52.0\\
    (PR’24) ICAFusion~\cite{DBLP:journals/pr/ShenCLZFY24}&YOLOv5&95.2&60.1\\
    Ours & YOLOv5&\textcolor{green}{97.6}&\textcolor{green}{65.2}\\
    \midrule
    YOLOv8l-IR~\cite{10533619}&YOLOv8&95.2&62.1\\
    YOLOv8l-RGB~\cite{10533619}&YOLOv8&91.9&54.0\\
    \midrule
    (ICHMS’24) MDMDet~\cite{10555757}&YOLOv7&96.5&61.5\\
    (RS'24) ACDF-YOLO~\cite{rs16183532}&YOLOv8&96.5&61.3\\
    (ICHMS'24) MSFF~\cite{10555757} &YOLOv8&96.8&61.9\\
    (Sensors'24) FAWDet~\cite{s24134098} &YOLOv8&97.1&62.1\\
    Ours & YOLOv8&\textcolor{red}{98.3}&\textcolor{red}{66.0}\\
    \bottomrule
  \end{tabular}}
  \caption{Comparison results with SOTA methods on LLVIP dataset. The best results are highlighted in \textcolor{red}{red}. The second and third best results are
highlighted in \textcolor{green}{green} and \textcolor{blue}{blue}, respectively.}
  \label{tab:example5}
\end{table}
\begin{table*}
  \centering
  \small
  \scalebox{0.75}{\begin{tabular}{cc|ccc|cc|cc}
    \toprule
Methods&Backbone&Precision&Recall&F1&$mAP_{50}$&$mAP$&Parameters&Inference time (ms)\\
    \midrule
    (TITS'23) MFPT~\cite{10105844}&ResNet50&79.3&72.9&76.0&80.0&41.9&200.0M&80.0\\
    Ours & ResNet50&\textbf{84.2}&\textbf{77.9}&\textbf{80.9}&\textbf{86.5}&\textbf{47.9}&\textbf{193.2M}&\textbf{53.2}\\
    \midrule
    (PRL’24) CrossFormer~\cite{LEE2024144}&YOLOv5&78.1&72.8&75.4&79.3&42.1&340.0M&80.0\\
    Ours & YOLOv5&\textbf{83.9}&\textbf{79.7}&\textbf{81.7}&\textbf{85.8}&\textbf{47.5}&\textbf{45.6M}&\textbf{44.1}\\
    \midrule
    YOLOv8-IR~\cite{10533619}&YOLOv8&75.1&65.3&69.9&72.9&38.3&43.7M&22\\
    YOLOv8-RGB~\cite{10533619}&YOLOv8&71.6&62.2&66.6&66.3&28.2&43.7M&22\\
    \midrule
    (GRSL'24) ESSFN~\cite{10653751}&YOLOv8&81.4&73.5&77.2&80.8&42.3&80.2M&47.0\\
    Ours & YOLOv8&\textbf{84.2}&\textbf{80.9}&\textbf{82.5}&\textbf{88.4}&\textbf{48.1}&\textbf{69.1M}&\textbf{40.0}\\
    \bottomrule
  \end{tabular}}
  \caption{Comparison results with SOTA methods on FLIR-Aligned dataset. The best results are highlighted in \textbf{bold}.}
  \label{tab:example6}
  \vspace{-10pt}
\end{table*}

\subsection{Experimental Setups}

We evaluate our model on six datasets with $mAP_{50}$ and $mAP$: $M^{3}$FD~\cite{9879642}, DroneVehicle~\cite{9759286}, LLVIP~\cite{9607632}, FLIR-Aligned~\cite{zhang2020multispectral} in the main paper and VEIDA~\cite{razakarivony2016vehicle}, KAIST~\cite{hwang2015multispectral} datasets in the supplementary.

Moreover, for the FLIR-Aligned dataset, we also employ precision, recall, and F1 score as evaluation metrics.

Our model and training framework are based on a modified dual-stream framework of the YOLO architecture by Ultralytics~\cite{10533619}. 
To verify the universality of our method, we implement our approach with three standard backbones: \textbf{ResNet50}, \textbf{YOLOv5} and \textbf{YOLOv8}, following existing methods~\cite{DBLP:journals/pr/ShenCLZFY24}.
The loss function and training pipeline are identical to those used in standard YOLOv8 training~\cite{10533619}.
More details can be found in the supplementary materials.

\subsection{Comparison with SOTA Results}
\noindent\textbf{$\mathbf{M^{3}FD}$ Dataset.} $M^{3}$FD is a multi-class object detection dataset in extreme weather conditions. The results on $M^{3}$FD are summarized in Table~\ref{tab:example3}. Compared to existing state-of-the-art methods, our YOLOv8-based method achieves the highest performance, with improvements of $5.5$\% and $5.1$\% in \( mAP_{50} \) and \( mAP \), respectively. 
Similarly, our method using ResNet50 and YOLOv5 as backbones also outperforms other methods using the same backbones. 
For performance in different categories, our method with different backbones consistently achieves top-three rankings in most cases.
Specifically, our method outperforms the fourth-ranked method by $1.1$\%, $6.0$\%, and $8.5$\% in the categories of cars, lamps, and people, respectively.
Note that even without special design for rare classes, our method still achieves similar performance in these rare classes (e.g., ``Bus'', ``Truck'', ``Motorcycle'').
Compared to the single-modal YOLOv8l model, our fusion models achieve significant performance improvements, demonstrating the effectiveness of our method in integrating beneficial information from both the IR and RGB modalities. 
As our method effectively leverages the global structural information from the low-frequency sub-band of IR and the detailed information, such as edge textures of objects, from the high-frequency sub-bands of RGB, we demonstrate exceptional detection performance under extreme weather conditions.
Furthermore, the consistent improvement across the three backbones indicates that our fusion approach is versatile and adaptable to various feature extraction methods, thus providing stable performance improvements. 
\\
\noindent\textbf{DroneVehicle Dataset.} 
DroneVehicle is a challenging remote sensing dataset with densely annotated images and small targets (e.g., ``car'', ``bus'', ``van''). As shown in Table~\ref{tab:example4}, our method achieves top-three rankings in both $mAP_{50}$ and $mAP$ metrics, surpassing the fourth-place method by $2.6$\% and $3.6$\%, respectively. Unlike other methods, which leverages domain-specific designs for remote sensing, our approach remains general yet still secures top-three positions in the categories of ``truck'' and ``freight-car''. Thanks to our unique low- and high-frequency fusion, our method excels at capturing edge details and achieves superior overall performance, highlighting its ability to enhance object detection across diverse scenarios without specialized remote sensing design.\\
\noindent\textbf{LLVIP Dataset.} 
LLVIP is a low-light dataset for pedestrian detection. 
As presented in Table~\ref{tab:example5}, we compare the performance of our approach on LLVIP dataset against state-of-the-art single-modal and fusion models. 
Our method achieves superior performance when using the same backbone.
In particular, our YOLOv8-based method achieves remarkable state-of-the-art results, with $98.3\%$ in $mAP_{50}$ and $66.0\%$ in $mAP$. \\
\noindent\textbf{FLIR-Aligned Dataset.} 
To demonstrate the advantages of our approach, we compare our methods with state-of-the-art methods with more metrics.
As shown in Table~\ref{tab:example6}, our method outperforms other fusion methods using the same backbone, achieving superior accuracy, while also reducing the size of the parameters and enhancing the speed of inference.
For example, compared to CrossFormer, our method surpasses $5.8\%$ in precision, $6.9\%$ in recall and $6.3\%$ in the F1 score, and reduces the model size by $294.4$ million parameters and decreases the inference time by $35.9$ milliseconds. 
MFPT and ESSFN exhibit a similar trend. 
This benefit is due to our fusion design and the linear time complexity of Mamba.
This further highlights the effectiveness of our approach in optimizing computational resources and improving detection performance. 
\subsection{Visualization}
\label{heatmaps}
\begin{figure}[tbp]
    \centering \includegraphics[width=0.45\textwidth]{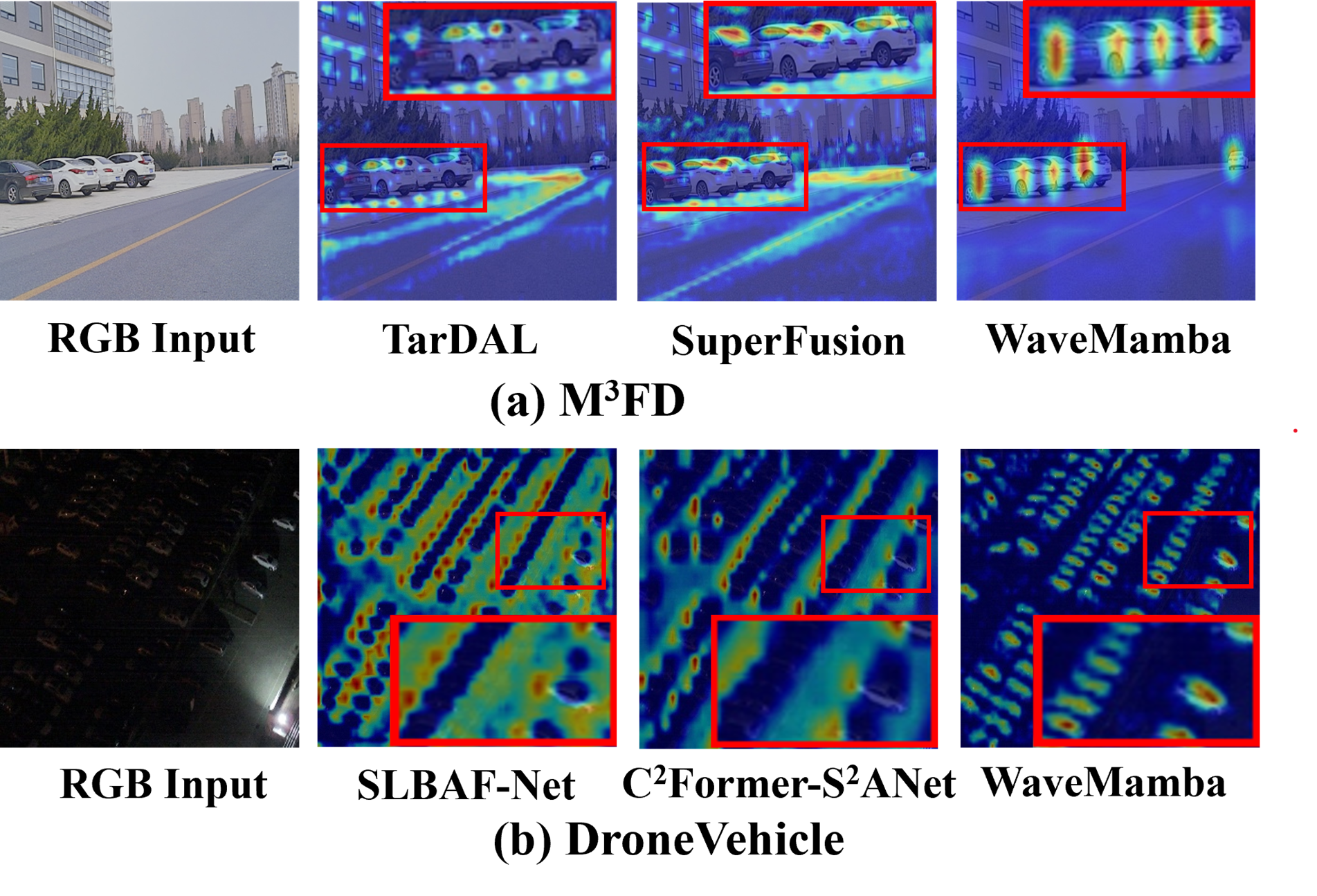}
    \caption{Heatmap visualization of several cross-modality object detection methods on $M^3$FD and DroneVehicle. } 
    \label{fig:placeholder2} 
    \vspace{-10pt}
\end{figure}
\begin{figure*}[tbp]
    \centering \includegraphics[width=0.85\linewidth]{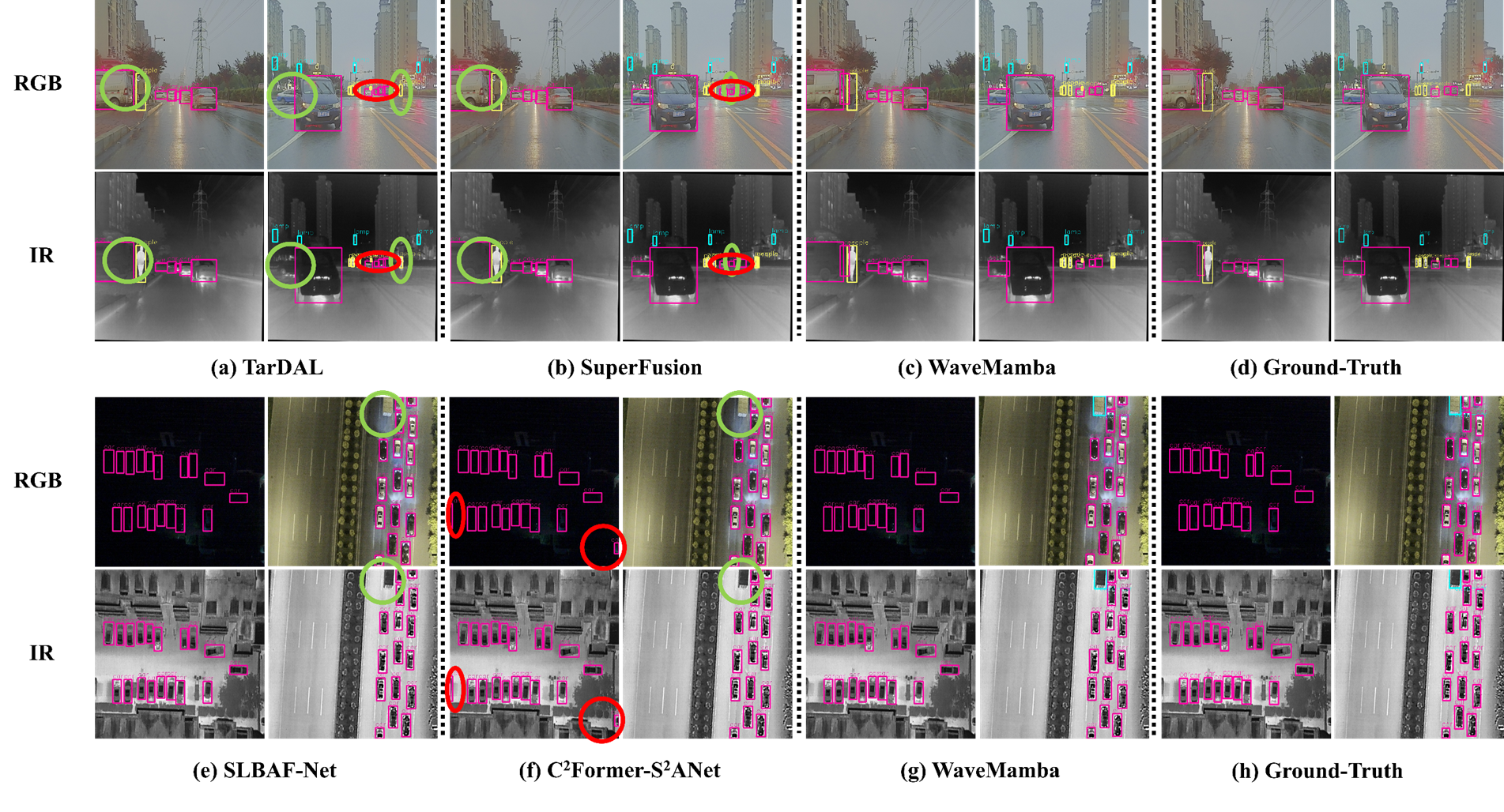}
    \caption{Detection results' visualization of several cross-modality object detection methods on $M^3$FD and DroneVehicle. Wherein, (a)-(d) present the results of $M^3$FD dataset, and (e)-(h) present the results of DroneVehicle dataset. The targets encircled by red ellipses are false positives, while those encircled by green ellipses are missed detections. Please zoom in for more details.} 
    \label{fig:placeholder3} 
    \vspace{-10pt}
\end{figure*}
\noindent\textbf{Heatmaps.}
We visualize the heatmap of our model's first inverse wavelet transform layer based on Grad-CAM~\cite{8237336} and compare it with other state-of-the-art methods~\cite{9879642,9970457,10.1007/s11042-023-15333-w,10472947}. 
As shown in Fig.~\ref{fig:placeholder2}, our model effectively prioritizes the target by utilizing wavelet transform and feature fusion, minimizing unnecessary attention on the background. 
In contrast, others tend to focus on larger areas, including backgrounds, which renders small targets indistinguishable.

\noindent\textbf{Detection Results.}~\label{detectionresults} 
We visualize our detection results and compare them with other state-of-the-art methods~\cite{9879642,9970457,10.1007/s11042-023-15333-w,10472947}. As shown in Fig.~\ref{fig:placeholder3},  our method reduces the number of missed or false detections in challenging scenarios (\ie, adverse weather, low lighting, heavy occlusion, or dense detection conditions), achieving the superior detection performance. 
\subsection{Ablation Study}
\label{Ablation study}
To verify the efficacy of each module and different fusion strategies, we perform ablation studies using the YOLOv8 backbone on the $M^{3}$FD dataset. More ablation studies such as the position and number of our WMFB are provided in the supplementary materials.

\noindent\textbf{Effects of the Improved Head.} 
We compare our improved head with the original YOLOv8 head in Table~\ref{tab:example7}.
As we can see, utilizing our improved head results in a $1.1$\% improvement in $mAP_{50}$ and $mAP$, coupled with a decrease of $7.6$M in the number of parameters.
The achievement can be attributed to the use of DWT and IDWT rather than down- and up-sampling. The use of DWT and IDWT preserves more detailed information, reduces information loss, and enhances the capture and reconstruction of features at multiple scales, resulting in improved overall model performance~\cite{8875369,10.1007/978-3-031-19806-9_19,finder2024waveletconvolutionslargereceptive}.

\noindent\textbf{Effects of the SFM and DFM Modules.}
In Table~\ref{tab:example8}, we ablate the impact of SFM and DFM Modules by removing these two modules separately.
Removing SFM leads to a decrease in performance by $1.5$\% in $mAP_{50}$ and $2.1$\% in $mAP$. Similarly, removing DFM results in a drop of $1.9$\% in $mAP_{50}$ and $2.2$\% in $mAP$.
This verifies the importance of SFM and DFM. Their fusion design, which progresses from shallow to deep levels, has exhibited a positive effect on the integration of features across various modalities.

\noindent\textbf{Effects of Different Fusion Strategies.} To verify the effectiveness of our proposed fusion strategies, we evaluate them against a standard fusion strategy Avg~\cite{9416456} in Table~\ref{tab:example10}. Here, (``A'', ``B'') indicates using strategy ``A'' and ``B'' to fuse high-frequency and low-frequency components, respectively. Starting with a simple baseline (Avg for fusion without DWT) and then progressively incorporating DWT along with different fusion strategies.The baseline lacks DWT and is incompatible with our improved YOLOv8 head, thus all experiments use the original head for fair comparison. By comparing the 1st and 2nd rows, we demonstrate the effectiveness of RGB-IR fusion in the wavelet domain, and further comparisons (2nd to 5th rows) show that our final combination (5th row) achieves the best performance, effectively preserving both low-frequency spatial information and high-frequency detail characteristics of objects.

\begin{table}
  \centering
  \scalebox{0.75}{
  \begin{tabular}{cccc}
    \toprule
    Methods&$mAP_{50}$ & $mAP$ & Parameters\\
    \midrule
    Origin YOLOv8 head& 91.0 &63.3 &76.7M\\
    Improved YOLOv8 head& 92.1&64.4 &69.1M\\
    \bottomrule
  \end{tabular}}
  \caption{Effects of the improved head on $M^{3}$FD dataset.}
  \label{tab:example7}
  \vspace{-10pt}
\end{table}

\begin{table}
  \centering
  \scalebox{0.75}{
  \begin{tabular}{cccc}
    \toprule
    Methods&$mAP_{50}$ & $mAP$ & Parameters\\
    \midrule
    w/o SFM module& 90.6 & 62.3 &63.4M\\
    w/o DFM module& 90.2 &62.2 &64.1M\\
    ours&92.1&64.4&69.1M\\
    \bottomrule
  \end{tabular}}
  \caption{Effects of the SFM and DFM  on $M^{3}$FD dataset.}
  \label{tab:example8}
  \vspace{-10pt}
\end{table}

\begin{table}
  \centering
  \scalebox{0.75}{
  \begin{tabular}{cccc}
    \toprule
    Methods&$mAP_{50}$ & $mAP$ & Parameters\\
    \midrule
    baseline& 83.2 & 55.1 & 68.9M\\
    (Avg, Avg)& 86.6 & 58.2 &66.3M\\
    (Avg, LMFB)& 90.6 &62.6 &76.7M\\
    (HFE, Avg)& 90.4 &62.3 &66.3M\\
    (HFE, LMFB) &91.0&63.3&76.7M\\
    \bottomrule
  \end{tabular}}
  \caption{Effects of different fusion strategies in WaveMamba Fusion Blocks on $M^{3}$FD dataset.}
  \label{tab:example10}
  \vspace{-10pt}
\end{table}

\section{Conclusion}
In this paper, we propose WaveMamba, a method that utilizes the unique and complementary  frequency characteristics of different modalities to integrate features, prompting extensive feature interaction and improving the effectiveness of RGB-IR object detection. 
The main innovation of WaveMamba lies in WMFB, which incorporates an LMFB to progressively integrate low-frequency features from shallow to deep layers, alongside a HFE strategy for efficient integration of high-frequency information.
Extensive experiments on four distinct types of public RGB-IR datasets demonstrate that our method
achieves state-of-the-art results while balancing parameters and efficiency.
As a result, this work establishes a new baseline for RGB-IR object detection, setting a precedent for future research in the field.
\section*{Acknowledgements}
The work was supported by the National Key Research and Development Program of China (Grant No. 2023YFC3306401). 
This research was also supported by the Zhejiang Provincial Natural Science Foundation of China (Grant No. LD24F020007), Beijing Natural Science Foundation (Grant No. L223024 and L244043), National Natural Science Foundation of China (Grant No. 62076016, 62406298 and 62176068), ``One Thousand Plan'' projects in Jiangxi Province Jxsg2023102268, Beijing Municipal Science $\&$ Technology Commission, Administrative Commission of Zhongguancun Science Park Grant No.Z231100005923035, Taiyuan City ``Double hundred Research action''  2024TYJB0127, the Fundamental Research Funds for the Central Universities (CUC25QT17).

{
    \small
    \bibliographystyle{ieeenat_fullname}
    \bibliography{main}

\begin{thebibliography}{103}
\providecommand{\natexlab}[1]{#1}
\providecommand{\url}[1]{\texttt{#1}}
\expandafter\ifx\csname urlstyle\endcsname\relax
  \providecommand{\doi}[1]{doi: #1}\else
  \providecommand{\doi}{doi: \begingroup \urlstyle{rm}\Url}\fi

\bibitem[Althoupety et~al.(2024)Althoupety, Wang, Feng, and Rekabdar]{10677953}
Afnan Althoupety, Li-Yun Wang, Wu-Chi Feng, and Banafsheh Rekabdar.
\newblock Daff: Dual attentive feature fusion for multispectral pedestrian detection.
\newblock In \emph{CVPRW}, pages 2997--3006, 2024.

\bibitem[Bao et~al.(2023)Bao, Cao, Hao, Cheng, Ning, and Zhao]{s23062934}
Chun Bao, Jie Cao, Qun Hao, Yang Cheng, Yaqian Ning, and Tianhua Zhao.
\newblock Dual-yolo architecture from infrared and visible images for object detection.
\newblock \emph{Sensors}, 2023.

\bibitem[Cao et~al.(2023)Cao, Bin, Hamari, Blasch, and Liu]{10209020}
Yue Cao, Junchi Bin, Jozsef Hamari, Erik Blasch, and Zheng Liu.
\newblock Multimodal object detection by channel switching and spatial attention.
\newblock In \emph{CVPRW}, 2023.

\bibitem[Chen et~al.(2024{\natexlab{a}})Chen, Qi, Liu, Bin, Fu, Hu, and Zhong]{10655285}
Chen Chen, Jiahao Qi, Xingyue Liu, Kangcheng Bin, Ruigang Fu, Xikun Hu, and Ping Zhong.
\newblock Weakly misalignment-free adaptive feature alignment for uavs-based multimodal object detection.
\newblock In \emph{CVPR}, 2024{\natexlab{a}}.

\bibitem[Chen et~al.(2024{\natexlab{b}})Chen, Wang, Zhu, and Yuan]{10598725}
Yishuo Chen, Boran Wang, Wenbin Zhu, and Jing Yuan.
\newblock Rgb-ir yolo combining modality-specific reconstruction and information integration.
\newblock In \emph{YAC}, 2024{\natexlab{b}}.

\bibitem[Chen et~al.(2022)Chen, Shi, Ye, Mertz, Ramanan, and Kong]{10.1007/978-3-031-20077-9_9}
Yi-Ting Chen, Jinghao Shi, Zelin Ye, Christoph Mertz, Deva Ramanan, and Shu Kong.
\newblock Multimodal object detection via-probabilistic ensembling.
\newblock In \emph{ECCV}, 2022.

\bibitem[Cheng et~al.(2023)Cheng, Geng, Wang, Wang, Sun, and Ding]{10.1007/s11042-023-15333-w}
Xiaolong Cheng, Keke Geng, Ziwei Wang, Jinhu Wang, Yuxiao Sun, and Pengbo Ding.
\newblock Slbaf-net: Super-lightweight bimodal adaptive fusion network for uav detection in low recognition environment.
\newblock \emph{Multimedia Tools Appl.}, 2023.

\bibitem[Devaguptapu et~al.(2019)Devaguptapu, Akolekar, M~Sharma, and N~Balasubramanian]{devaguptapu2019borrow}
Chaitanya Devaguptapu, Ninad Akolekar, Manuj M~Sharma, and Vineeth N~Balasubramanian.
\newblock Borrow from anywhere: Pseudo multi-modal object detection in thermal imagery.
\newblock In \emph{CVPRW}, pages 0--0, 2019.

\bibitem[Fang et~al.(2021)Fang, Han, and Wang]{DBLP:journals/corr/abs-2111-00273}
Qingyun Fang, Dapeng Han, and Zhaokui Wang.
\newblock Cross-modality fusion transformer for multispectral object detection.
\newblock \emph{arXiv:2111.00273}, 2021.

\bibitem[Fei et~al.(2024)Fei, Guo, Li, Yu, and Sun]{rs16183532}
Xuan Fei, Mengyao Guo, Yan Li, Renping Yu, and Le Sun.
\newblock Acdf-yolo: Attentive and cross-differential fusion network for multimodal remote. sens. object detection.
\newblock \emph{Remote. Sens.}, 2024.

\bibitem[Finder et~al.(2024)Finder, Amoyal, Treister, and Freifeld]{finder2024waveletconvolutionslargereceptive}
Shahaf~E. Finder, Roy Amoyal, Eran Treister, and Oren Freifeld.
\newblock Wavelet convolutions for large receptive fields.
\newblock \emph{arXiv:2407.05848}, 2024.

\bibitem[FLIR(2024)]{reference1}
TELEDYNE FLIR.
\newblock Free teledyne flir thermal dataset for algorithm training.
\newblock Online, 2024.

\bibitem[Fu et~al.(2024)Fu, Wang, Duan, Xiao, Dian, Li, and Li]{10144688}
Haolong Fu, Shixun Wang, Puhong Duan, Changyan Xiao, Renwei Dian, Shutao Li, and Zhiyong Li.
\newblock Lraf-net: Long-range attention fusion network for visible–infrared object detection.
\newblock \emph{TNNLS}, 35\penalty0 (10):\penalty0 13232--13245, 2024.

\bibitem[Gao et~al.(2024)Gao, Kanu{-}Asiegbu, and Du]{MambaST}
Xiangbo Gao, Asiegbu~Miracle Kanu{-}Asiegbu, and Xiaoxiao Du.
\newblock Mambast: {A} plug-and-play cross-spectral spatial-temporal fuser for efficient pedestrian detection.
\newblock \emph{arXiv:2408.01037}, 2024.

\bibitem[Gu and Dao(2023)]{gu2023mamba}
Albert Gu and Tri Dao.
\newblock Mamba: Linear-time sequence modeling with selective state spaces.
\newblock \emph{arXiv:2312.00752}, 2023.

\bibitem[Guo et~al.(2024)Guo, Gao, Liu, Meng, and Gao]{guo2024damsdet}
Junjie Guo, Chenqiang Gao, Fangcen Liu, Deyu Meng, and Xinbo Gao.
\newblock Damsdet: Dynamic adaptive multispectral detection transformer with competitive query selection and adaptive feature fusion.
\newblock In \emph{ECCV}, pages 464--481, 2024.

\bibitem[He et~al.(2024)He, Cao, Zhang, Yan, Wang, Li, Xie, Hong, and Zhou]{Panmamba}
Xuanhua He, Ke Cao, Jie Zhang, Keyu Yan, Yingying Wang, Rui Li, Chengjun Xie, Danfeng Hong, and Man Zhou.
\newblock Pan-mamba: Effective pan-sharpening with state space model.
\newblock \emph{CoRR'24}, 2024.

\bibitem[Huang et~al.(2022)Huang, Liu, Wang, Yuan, and Chen]{HUANG2022104737}
Lida Huang, Gang Liu, Yan Wang, Hongyong Yuan, and Tao Chen.
\newblock Fire detection in video surveillances using convolutional neural networks and wavelet transform.
\newblock \emph{EAAI}, 110:\penalty0 104737, 2022.

\bibitem[Huang et~al.(2024)Huang, Peng, Chen, Dai, He, and Liu]{HUANG2024106156}
Lian Huang, Zongju Peng, Fen Chen, Shaosheng Dai, Ziqiang He, and Kesheng Liu.
\newblock Cross-modality interaction for few-shot multispectral object detection with semantic knowledge.
\newblock \emph{Neural Networks}, 2024.

\bibitem[Hwang et~al.(2015)Hwang, Park, Kim, Choi, and So~Kweon]{hwang2015multispectral}
Soonmin Hwang, Jaesik Park, Namil Kim, Yukyung Choi, and In So~Kweon.
\newblock Multispectral pedestrian detection: Benchmark dataset and baseline.
\newblock In \emph{CVPR}, pages 1037--1045, 2015.

\bibitem[Jia et~al.(2021)Jia, Zhu, Li, Tang, and Zhou]{9607632}
Xinyu Jia, Chuang Zhu, Minzhen Li, Wenqi Tang, and Wenli Zhou.
\newblock Llvip: A visible-infrared paired dataset for low-light vision.
\newblock In \emph{ICCVW}, 2021.

\bibitem[Jiang et~al.(2024)Jiang, Ren, Yang, Huo, Zhu, Yao, Li, Sun, and Yang]{JIANG2024103918}
Chenchen Jiang, Huazhong Ren, Hong Yang, Hongtao Huo, Pengfei Zhu, Zhaoyuan Yao, Jing Li, Min Sun, and Shihao Yang.
\newblock M2fnet: Multi-modal fusion network for object detection from visible and thermal infrared images.
\newblock \emph{Int. J. Appl. Earth Obs. Geoinf.}, 2024.

\bibitem[Jie et~al.(2024)Jie, Xu, Li, and Tan]{jie2024tsjnetmultimodalitytargetsemantic}
Yuchan Jie, Yushen Xu, Xiaosong Li, and Haishu Tan.
\newblock Tsjnet: A multi-modality target and semantic awareness joint-driven image fusion network.
\newblock \emph{arXiv:2402.01212}, 2024.

\bibitem[Kang et~al.(2024)Kang, Yin, and Duan]{10476333}
Xudong Kang, Hui Yin, and Puhong Duan.
\newblock Global–local feature fusion network for visible–infrared vehicle detection.
\newblock \emph{IEEE Geosci. Remote Sens. Lett.}, 2024.

\bibitem[Kim et~al.(2022)Kim, Park, and Ro]{KimPR22}
Jung~Uk Kim, Sungjune Park, and Yong~Man Ro.
\newblock Uncertainty-guided cross-modal learning for robust multispectral pedestrian detection.
\newblock \emph{TCSVT}, 2022.

\bibitem[Lee et~al.(2024)Lee, Park, and Park]{LEE2024144}
Seungik Lee, Jaehyeong Park, and Jinsun Park.
\newblock Crossformer: Cross-guided attention for multi-modal object detection.
\newblock \emph{Pattern Recognit. Lett.}, 2024.

\bibitem[Lee et~al.(2023)Lee, Jovanov, and Philips]{10.1007/978-3-031-25072-9_41}
Wei-Yu Lee, Ljubomir Jovanov, and Wilfried Philips.
\newblock Cross-modality attention and multimodal fusion transformer for pedestrian detection.
\newblock In \emph{ECCVW}, 2023.

\bibitem[Li et~al.(2024{\natexlab{a}})Li, Wang, Wang, Liu, Yin, Fang, and Geng]{10648743}
Ang Li, Ziwei Wang, Fanxun Wang, Zhichao Liu, Guodong Yin, Ruiqi Fang, and Keke Geng.
\newblock A novel semantic information perception architecture for extreme targets detection in complex traffic scenarios.
\newblock \emph{IEEE Trans. Intell. Veh.}, 2024{\natexlab{a}}.

\bibitem[Li et~al.(2019)Li, Song, Tong, and Tang]{LI2019161}
Chengyang Li, Dan Song, Ruofeng Tong, and Min Tang.
\newblock Illumination-aware faster r-cnn for robust multispectral pedestrian detection.
\newblock \emph{PR}, 2019.

\bibitem[Li et~al.(2023{\natexlab{a}})Li, Chen, Liu, and Ma]{10.1145/3581783.3612135}
Jiawei Li, Jiansheng Chen, Jinyuan Liu, and Huimin Ma.
\newblock Learning a graph neural network with cross modality interaction for image fusion.
\newblock In \emph{ACMMM}, 2023{\natexlab{a}}.

\bibitem[Li et~al.(2023{\natexlab{b}})Li, Zhang, Hu, Fu, and Zhu]{9739079}
Qing Li, Changqing Zhang, Qinghua Hu, Huazhu Fu, and Pengfei Zhu.
\newblock Confidence-aware fusion using dempster-shafer theory for multispectral pedestrian detection.
\newblock \emph{IEEE Trans. Multimed.}, 2023{\natexlab{b}}.

\bibitem[Li et~al.(2024{\natexlab{b}})Li, Zhang, Hu, Zhu, Fu, and Chen]{10225383}
Qing Li, Changqing Zhang, Qinghua Hu, Pengfei Zhu, Huazhu Fu, and Lei Chen.
\newblock Stabilizing multispectral pedestrian detection with evidential hybrid fusion.
\newblock \emph{TCSVT}, 2024{\natexlab{b}}.

\bibitem[Li et~al.(2024{\natexlab{c}})Li, Xiang, Sun, Yuan, Yuan, and Gou]{10114594}
Ruimin Li, Jiajun Xiang, Feixiang Sun, Ye Yuan, Longwu Yuan, and Shuiping Gou.
\newblock Multiscale cross-modal homogeneity enhancement and confidence-aware fusion for multispectral pedestrian detection.
\newblock \emph{IEEE Trans. Multimed.}, 2024{\natexlab{c}}.

\bibitem[Li et~al.(2024{\natexlab{d}})Li, Chen, Tian, Zhou, and Zhang]{10478590}
Xiangyang Li, Shiguo Chen, Chunna Tian, Heng Zhou, and Zhenxi Zhang.
\newblock M2fnet: Mask-guided multi-level fusion for rgb-t pedestrian detection.
\newblock \emph{IEEE Trans. Multimed.}, 2024{\natexlab{d}}.

\bibitem[Liang et~al.(2023)Liang, Hu, Bao, Feng, Deng, and Lam]{Liang2023ExplicitAF}
Mingjian Liang, Junjie Hu, Chenyu Bao, Hua Feng, Fuqin Deng, and Tin~Lun Lam.
\newblock Explicit attention-enhanced fusion for rgb-thermal perception tasks.
\newblock \emph{IEEE Robot. Autom. Lett.}, 2023.

\bibitem[Liu et~al.(2016{\natexlab{a}})Liu, Zhang, Wang, and Metaxas]{LiuZWM16}
Jingjing Liu, Shaoting Zhang, Shu Wang, and Dimitris~N. Metaxas.
\newblock Multispectral deep neural networks for pedestrian detection.
\newblock In \emph{BMVC}, 2016{\natexlab{a}}.

\bibitem[Liu et~al.(2016{\natexlab{b}})Liu, Zhang, Wang, and Metaxas]{liu2016multispectral}
Jingjing Liu, Shaoting Zhang, Shu Wang, and Dimitris~N Metaxas.
\newblock Multispectral deep neural networks for pedestrian detection.
\newblock \emph{arXiv preprint arXiv:1611.02644}, 2016{\natexlab{b}}.

\bibitem[Liu et~al.(2022)Liu, Fan, Huang, Wu, Liu, Zhong, and Luo]{9879642}
Jinyuan Liu, Xin Fan, Zhanbo Huang, Guanyao Wu, Risheng Liu, Wei Zhong, and Zhongxuan Luo.
\newblock Target-aware dual adversarial learning and a multi-scenario multi-modality benchmark to fuse infrared and visible for object detection.
\newblock In \emph{CVPR}, 2022.

\bibitem[Liu et~al.(2017)Liu, Chen, Peng, and Wang]{LIU2017191}
Yu Liu, Xun Chen, Hu Peng, and Zengfu Wang.
\newblock Multi-focus image fusion with a deep convolutional neural network.
\newblock \emph{Inf. Fusion}, 36:\penalty0 191--207, 2017.

\bibitem[Liu et~al.(2024)Liu, Tian, Zhao, Yu, Xie, Wang, Ye, and Liu]{liu2024vmamba}
Yue Liu, Yunjie Tian, Yuzhong Zhao, Hongtian Yu, Lingxi Xie, Yaowei Wang, Qixiang Ye, and Yunfan Liu.
\newblock Vmamba: Visual state space model.
\newblock \emph{arXiv preprint arXiv:2401.10166}, 2024.

\bibitem[Ma et~al.(2024{\natexlab{a}})Ma, Li, Cheng, Wang, Song, and Wu]{ma2024s4fusionsaliencyawareselectivestate}
Haolong Ma, Hui Li, Chunyang Cheng, Gaoang Wang, Xiaoning Song, and Xiaojun Wu.
\newblock S4fusion: Saliency-aware selective state space model for infrared visible image fusion, 2024{\natexlab{a}}.

\bibitem[Ma et~al.(2024{\natexlab{b}})Ma, Ni, and Chen]{ma2024tinyvim}
Xiaowen Ma, Zhenliang Ni, and Xinghao Chen.
\newblock Tinyvim: Frequency decoupling for tiny hybrid vision mamba.
\newblock \emph{arXiv preprint arXiv:2411.17473}, 2024{\natexlab{b}}.

\bibitem[Nie et~al.(2024)Nie, Sun, Sun, Ni, and Gao]{10342625}
Jinyan Nie, He Sun, Xu Sun, Li Ni, and Lianru Gao.
\newblock Cross-modal feature fusion and interaction strategy for cnn-transformer-based object detection in visual and infrared remote sensing imagery.
\newblock \emph{IEEE Geoscience and Remote. Sens. Letters}, 21:\penalty0 1--5, 2024.

\bibitem[Qingyun and Zhaokui(2022)]{QINGYUN2022108786}
Fang Qingyun and Wang Zhaokui.
\newblock Cross-modality attentive feature fusion for object detection in multispectral remote. sens. imagery.
\newblock \emph{PR}, 2022.

\bibitem[Qingyun et~al.(2021)Qingyun, Dapeng, and Zhaokui]{qingyun2021cross}
Fang Qingyun, Han Dapeng, and Wang Zhaokui.
\newblock Cross-modality fusion transformer for multispectral object detection.
\newblock \emph{arXiv preprint arXiv:2111.00273}, 2021.

\bibitem[Razakarivony and Jurie(2016)]{razakarivony2016vehicle}
Sebastien Razakarivony and Frederic Jurie.
\newblock Vehicle detection in aerial imagery: A small target detection benchmark.
\newblock \emph{JVCIR}, 34:\penalty0 187--203, 2016.

\bibitem[Selvaraju et~al.(2017)Selvaraju, Cogswell, Das, Vedantam, Parikh, and Batra]{8237336}
Ramprasaath~R. Selvaraju, Michael Cogswell, Abhishek Das, Ramakrishna Vedantam, Devi Parikh, and Dhruv Batra.
\newblock Grad-cam: Visual explanations from deep networks via gradient-based localization.
\newblock In \emph{ICCV}, pages 618--626, 2017.

\bibitem[Shen et~al.(2024)Shen, Chen, Liu, Zuo, Fan, and Yang]{DBLP:journals/pr/ShenCLZFY24}
Jifeng Shen, Yifei Chen, Yue Liu, Xin Zuo, Heng Fan, and Wankou Yang.
\newblock Icafusion: Iterative cross-attention guided feature fusion for multispectral object detection.
\newblock \emph{PR}, 2024.

\bibitem[Sun et~al.(2024{\natexlab{a}})Sun, Chen, Qiu, Li, and You]{s24103222}
Chaoyue Sun, Yajun Chen, Xiaoyang Qiu, Rongzhen Li, and Longxiang You.
\newblock Mrd-yolo: A multispectral object detection algorithm for complex road scenes.
\newblock \emph{Sensors}, 2024{\natexlab{a}}.

\bibitem[Sun et~al.(2024{\natexlab{b}})Sun, Yu, and Cheng]{10643097}
Xu Sun, Yinhui Yu, and Qing Cheng.
\newblock Low-rank multimodal remote. sens. object detection with frequency filtering experts.
\newblock \emph{IEEE Trans. Geosci. Remote Sens.}, 2024{\natexlab{b}}.

\bibitem[Sun et~al.(2024{\natexlab{c}})Sun, Yu, and Cheng]{article}
Xu Sun, Yinhui Yu, and Qing Cheng.
\newblock Adaptive multimodal feature fusion with frequency domain gate for remote. sens. object detection.
\newblock \emph{Remote Sens. Lett.}, 2024{\natexlab{c}}.

\bibitem[Sun et~al.(2022{\natexlab{a}})Sun, Cao, Zhu, and Hu]{10.1145/3503161.3547902}
Yiming Sun, Bing Cao, Pengfei Zhu, and Qinghua Hu.
\newblock Detfusion: A detection-driven infrared and visible image fusion network.
\newblock In \emph{ACMMM}, page 4003–4011, 2022{\natexlab{a}}.

\bibitem[Sun et~al.(2022{\natexlab{b}})Sun, Cao, Zhu, and Hu]{9759286}
Yiming Sun, Bing Cao, Pengfei Zhu, and Qinghua Hu.
\newblock Drone-based rgb-infrared cross-modality vehicle detection via uncertainty-aware learning.
\newblock \emph{TCSVT}, 2022{\natexlab{b}}.

\bibitem[Tang et~al.(2022)Tang, Deng, Ma, Huang, and Ma]{9970457}
Linfeng Tang, Yuxin Deng, Yong Ma, Jun Huang, and Jiayi Ma.
\newblock Superfusion: A versatile image registration and fusion network with semantic awareness.
\newblock \emph{IEEE/CAA J. Autom. Sin.}, 2022.

\bibitem[Tang et~al.(2023)Tang, Xiang, Zhang, Gong, and Ma]{TANG2023477}
Linfeng Tang, Xinyu Xiang, Hao Zhang, Meiqi Gong, and Jiayi Ma.
\newblock Divfusion: Darkness-free infrared and visible image fusion.
\newblock \emph{Inf. Fusion}, 2023.

\bibitem[Tian et~al.(2023)Tian, Zheng, Zuo, Zhang, Zhang, and Zhang]{TIAN2023109050}
Chunwei Tian, Menghua Zheng, Wangmeng Zuo, Bob Zhang, Yanning Zhang, and David Zhang.
\newblock Multi-stage image denoising with the wavelet transform.
\newblock \emph{PR}, 134:\penalty0 109050, 2023.

\bibitem[Tian et~al.(2024{\natexlab{a}})Tian, Zhou, Huang, Li, and He]{10382506}
Chao Tian, Zikun Zhou, Yuqing Huang, Gaojun Li, and Zhenyu He.
\newblock Cross-modality proposal-guided feature mining for unregistered rgb-thermal pedestrian detection.
\newblock \emph{IEEE Trans. Multimed.}, 2024{\natexlab{a}}.

\bibitem[Tian et~al.(2024{\natexlab{b}})Tian, Yan, Zhou, Wang, and Zhang]{s24196181}
Dan Tian, Xin Yan, Dong Zhou, Chen Wang, and Wenshuai Zhang.
\newblock Iv-yolo: A lightweight dual-branch object detection network.
\newblock \emph{Sensors}, 2024{\natexlab{b}}.

\bibitem[Tian et~al.(2025)Tian, Wang, Cao, Kang, Sun, Tian, Xing, Shen, Fan, Du, Fu, and Zhang]{10797700}
Shu Tian, Li Wang, Lin Cao, Lihong Kang, Xian Sun, Jing Tian, Xiangwei Xing, Bo Shen, Chunzhuo Fan, Kangning Du, Chong Fu, and Ye Zhang.
\newblock A dynamic cascade cross-modal coassisted network for aav image object detection.
\newblock \emph{J-STARS}, 18:\penalty0 2749--2765, 2025.

\bibitem[Varghese and M.(2024)]{10533619}
Rejin Varghese and Sambath M.
\newblock Yolov8: A novel object detection algorithm with enhanced performance and robustness.
\newblock In \emph{ADICS}, 2024.

\bibitem[Wang et~al.(2024{\natexlab{a}})Wang, Qu, Qiao, and Liu]{10632179}
Haoyu Wang, Shiyuan Qu, Zhenzhuang Qiao, and Xiaomin Liu.
\newblock Kcdnet: Multimodal object detection in modal information imbalance scenes.
\newblock \emph{{IEEE} Trans. Instrum. Meas.}, 2024{\natexlab{a}}.

\bibitem[Wang et~al.(2024{\natexlab{b}})Wang, Wang, Fu, Si, Zhang, Kou, Yu, and Feng]{10643643}
Huiying Wang, Chunping Wang, Qiang Fu, Binqiang Si, Dongdong Zhang, Renke Kou, Ying Yu, and Changfeng Feng.
\newblock Yolofiv: Object detection algorithm for around-the-clock aerial remote. sens. images by fusing infrared and visible features.
\newblock \emph{IEEE J. Sel. Top. Appl. Earth Obs. Remote Sens.}, 2024{\natexlab{b}}.

\bibitem[Wang et~al.(2024{\natexlab{c}})Wang, Wang, Fu, Zhang, Kou, Yu, and Song]{10440361}
Huiying Wang, Chunping Wang, Qiang Fu, Dongdong Zhang, Renke Kou, Ying Yu, and Jian Song.
\newblock Cross-modal oriented object detection of uav aerial images based on image feature.
\newblock \emph{IEEE Trans. Geosci. Remote Sens.}, 2024{\natexlab{c}}.

\bibitem[Wang et~al.(2024{\natexlab{d}})Wang, Su, Zhao, Yan, and Feng]{rs16203904}
Jinpeng Wang, Nan Su, Chunhui Zhao, Yiming Yan, and Shou Feng.
\newblock Multi-modal object detection method based on dual-branch asymmetric attention backbone and feature fusion pyramid network.
\newblock \emph{Remote. Sens.}, 2024{\natexlab{d}}.

\bibitem[Wang et~al.(2024{\natexlab{e}})Wang, Xu, Zhao, Gao, Wu, Yan, Feng, and Su]{10461034}
Jinpeng Wang, Congan Xu, Chunhui Zhao, Long Gao, Junfeng Wu, Yiming Yan, Shou Feng, and Nan Su.
\newblock Multimodal object detection of uav remote sensing based on joint representation optimization and specific information enhancement.
\newblock \emph{J-STARS}, 17:\penalty0 12364--12373, 2024{\natexlab{e}}.

\bibitem[Wang et~al.(2024{\natexlab{f}})Wang, Tu, Li, Zhang, and Luo]{WangTLZL24}
Kunpeng Wang, Zhengzheng Tu, Chenglong Li, Cheng Zhang, and Bin Luo.
\newblock Learning adaptive fusion bank for multi-modal salient object detection.
\newblock \emph{TCSVT}, 2024{\natexlab{f}}.

\bibitem[Wang et~al.(2022{\natexlab{a}})Wang, Chi, Shen, Song, Zhang, and Zhu]{rs14092020}
Qingwang Wang, Yongke Chi, Tao Shen, Jian Song, Zifeng Zhang, and Yan Zhu.
\newblock Improving rgb-infrared object detection by reducing cross-modality redundancy.
\newblock \emph{Remote. Sens.}, 14, 2022{\natexlab{a}}.

\bibitem[Wang et~al.(2022{\natexlab{b}})Wang, Chi, Shen, Song, Zhang, and Zhu]{wang2022improving}
Qingwang Wang, Yongke Chi, Tao Shen, Jian Song, Zifeng Zhang, and Yan Zhu.
\newblock Improving rgb-infrared object detection by reducing cross-modality redundancy.
\newblock \emph{Remote. Sens.}, 14\penalty0 (9):\penalty0 2020, 2022{\natexlab{b}}.

\bibitem[Wang et~al.(2022{\natexlab{c}})Wang, Zhuang, Xu, Ye, Xiao, and Peng]{wang2022imagerestorationqualityassessment}
Zhiyu Wang, Jiayan Zhuang, Ningyuan Xu, Sichao Ye, Jiangjian Xiao, and Chengbin Peng.
\newblock Image restoration quality assessment based on regional differential information entropy.
\newblock \emph{arxiv2107.03642}, 2022{\natexlab{c}}.

\bibitem[Wu et~al.(2023)Wu, Han, Yang, Zhao, Rao, Li, Xing, Zhou, and Bai]{DBLP:journals/tim/WuHYZRLXZB23}
Dan Wu, Mina Han, Yang Yang, Shan Zhao, Yujing Rao, Hao Li, Lin Xing, Chengjiang Zhou, and Haicheng Bai.
\newblock Dcfusion: {A} dual-frequency cross-enhanced fusion network for infrared and visible image fusion.
\newblock \emph{{IEEE} Trans. Instrum. Meas.}, 2023.

\bibitem[Wu et~al.(2024)Wu, Wang, Wang, Wang, and Gao]{DBLP:journals/sensors/WuWWWG24}
Dan Wu, Yanzhi Wang, Haoran Wang, Fei Wang, and Guowang Gao.
\newblock Dcfnet: Infrared and visible image fusion network based on discrete wavelet transform and convolutional neural network.
\newblock \emph{Sensors}, 2024.

\bibitem[Xiao et~al.(2019)Xiao, Wang, and Liu]{8875369}
Guangrun Xiao, Xiaobo Wang, and Dezheng Liu.
\newblock Wavelet transformation of functional data for hyperspectral image classification.
\newblock In \emph{2019 International Conference on Internet of Things (iThings) and IEEE Green Computing and Communications (GreenCom) and IEEE Cyber, Physical and Social Computing (CPSCom) and IEEE Smart Data (SmartData)}, pages 403--409, 2019.

\bibitem[Xie et~al.(2023{\natexlab{a}})Xie, Zhang, Yu, and Xie]{DBLP:journals/tamd/XieZYX23}
Yumin Xie, Langwen Zhang, Xiaoyuan Yu, and Wei Xie.
\newblock {YOLO-MS:} multispectral object detection via feature interaction and self-attention guided fusion.
\newblock \emph{IEEE Trans. Cogn. Dev. Syst.}, 2023{\natexlab{a}}.

\bibitem[Xie et~al.(2023{\natexlab{b}})Xie, Shao, Chen, Chen, Jiang, Meng, and Ho]{XieSCCJMH23}
Zhengxuan Xie, Feng Shao, Gang Chen, Hangwei Chen, Qiuping Jiang, Xiangchao Meng, and Yo{-}Sung Ho.
\newblock Cross-modality double bidirectional interaction and fusion network for {RGB-T} salient object detection.
\newblock \emph{TCSVT}, 2023{\natexlab{b}}.

\bibitem[Xu et~al.(2024{\natexlab{a}})Xu, Xu, Hong, Peng, Guo, and Li]{10653751}
Fengxiang Xu, Tingfa Xu, Lang Hong, Peiran Peng, Jiaxin Guo, and Jianan Li.
\newblock Enhanced spectral–spatial fusion network for multispectral object detection in ground-aerial images.
\newblock \emph{IEEE Geoscience and Remote. Sens. Letters}, 2024{\natexlab{a}}.

\bibitem[Xu et~al.(2024{\natexlab{b}})Xu, Xu, Hong, Peng, Guo, and Li]{xu2024enhanced}
Fengxiang Xu, Tingfa Xu, Lang Hong, Peiran Peng, Jiaxin Guo, and Jianan Li.
\newblock Enhanced spectral-spatial fusion network for multispectral object detection in ground-aerial images.
\newblock \emph{IEEE Geoscience and Remote. Sens. Letters}, 2024{\natexlab{b}}.

\bibitem[Xu et~al.(2022)Xu, Ma, Yuan, Le, and Liu]{9878923}
Han Xu, Jiayi Ma, Jiteng Yuan, Zhuliang Le, and Wei Liu.
\newblock Rfnet: Unsupervised network for mutually reinforcing multi-modal image registration and fusion.
\newblock In \emph{CVPR}, 2022.

\bibitem[Xu et~al.(2024{\natexlab{c}})Xu, Mo, Zhao, Zhao, Tao, and Han]{10598791}
Junwei Xu, Bo Mo, Jie Zhao, Chunbo Zhao, Yimeng Tao, and Shuo Han.
\newblock Cross-modal adaptive fusion object detection based on illumination-awareness.
\newblock In \emph{YAC}, 2024{\natexlab{c}}.

\bibitem[Yang et~al.(2024{\natexlab{a}})Yang, Liang, Li, and Zhang]{10666754}
Fan Yang, Binbin Liang, Wei Li, and Jianwei Zhang.
\newblock Multidimensional fusion network for multispectral object detection.
\newblock \emph{TCSVT}, 2024{\natexlab{a}}.

\bibitem[Yang et~al.(2024{\natexlab{b}})Yang, Bian, Wang, Bo, and Mi]{10555757}
Huanyu Yang, Weiwei Bian, Jun Wang, Yuming Bo, and Ying Mi.
\newblock A dual-modality pedestrian detection method based on multi-scale feature fusion.
\newblock In \emph{ICHMS}, 2024{\natexlab{b}}.

\bibitem[Yang et~al.(2024{\natexlab{c}})Yang, Yang, Liao, Huang, He, Zhang, Zhou, and Song]{10.1117/12.3040116}
Jinqi Yang, Xin Yang, Yizhao Liao, Jinxiang Huang, Hongyu He, Erfan Zhang, Ya Zhou, and Yong Song.
\newblock {Multispectral sample augmentation and illumination guidance for RGB-T object detection by mm detection framework}.
\newblock In \emph{LOPET}, 2024{\natexlab{c}}.

\bibitem[Yao et~al.(2022)Yao, Pan, Li, Ngo, and Mei]{10.1007/978-3-031-19806-9_19}
Ting Yao, Yingwei Pan, Yehao Li, Chong-Wah Ngo, and Tao Mei.
\newblock Wave-vit: Unifying wavelet and transformers for visual representation learning.
\newblock In \emph{ECCV}, 2022.

\bibitem[Yi et~al.(2024)Yi, Xu, Zhang, Tang, and Ma]{Yi_2024_CVPR}
Xunpeng Yi, Han Xu, Hao Zhang, Linfeng Tang, and Jiayi Ma.
\newblock Text-if: Leveraging semantic text guidance for degradation-aware and interactive image fusion.
\newblock In \emph{CVPR}, 2024.

\bibitem[Yuan and Wei(2024)]{10472947}
Maoxun Yuan and Xingxing Wei.
\newblock C²former: Calibrated and complementary transformer for rgb-infrared object detection.
\newblock \emph{IEEE Trans. Geosci. Remote Sens.}, 2024.

\bibitem[Yuan et~al.(2022)Yuan, Wang, and Wei]{10.1007/978-3-031-20077-9_30}
Maoxun Yuan, Yinyan Wang, and Xingxing Wei.
\newblock Translation, scale and rotation: Cross-modal alignment meets rgb-infrared vehicle detection.
\newblock In \emph{ECCV}, 2022.

\bibitem[Zhang and Ma(2021)]{Zhang2021SDNetAV}
Hao Zhang and Jiayi Ma.
\newblock Sdnet: A versatile squeeze-and-decomposition network for real-time image fusion.
\newblock \emph{IJCV}, 2021.

\bibitem[Zhang et~al.(2020)Zhang, Fromont, Lef{\`{e}}vre, and Avignon]{zhang2020multispectral}
Heng Zhang, {\'{E}}lisa Fromont, S{\'{e}}bastien Lef{\`{e}}vre, and Bruno Avignon.
\newblock Multispectral fusion for object detection with cyclic fuse-and-refine blocks.
\newblock In \emph{ICIP}, 2020.

\bibitem[Zhang et~al.(2021)Zhang, Fromont, Lefevre, and Avignon]{Zhang_2021_WACV}
Heng Zhang, Elisa Fromont, Sebastien Lefevre, and Bruno Avignon.
\newblock Guided attentive feature fusion for multispectral pedestrian detection.
\newblock In \emph{WACV}, pages 72--80, 2021.

\bibitem[Zhang et~al.(2024{\natexlab{a}})Zhang, qing Chang, yang Zhao, dun Ma, Han, and Zhang]{ZHANG2024111971}
Jie Zhang, Tian qing Chang, Li yang Zhao, Jin dun Ma, Bin Han, and Lei Zhang.
\newblock Efficient cross-modality feature interaction for multispectral armored vehicle detection.
\newblock \emph{Appl. Soft Comput}, 163:\penalty0 111971, 2024{\natexlab{a}}.

\bibitem[Zhang et~al.(2019{\natexlab{a}})Zhang, Liu, Zhang, Yang, Qiao, Huang, and Hussain]{ZHANG201920}
Lu Zhang, Zhiyong Liu, Shifeng Zhang, Xu Yang, Hong Qiao, Kaizhu Huang, and Amir Hussain.
\newblock Cross-modality interactive attention network for multispectral pedestrian detection.
\newblock \emph{Inf. Fusion}, 2019{\natexlab{a}}.

\bibitem[Zhang et~al.(2019{\natexlab{b}})Zhang, Zhu, Chen, Yang, Lei, and Liu]{zhang2019weakly}
Lu Zhang, Xiangyu Zhu, Xiangyu Chen, Xu Yang, Zhen Lei, and Zhiyong Liu.
\newblock Weakly aligned cross-modal learning for multispectral pedestrian detection.
\newblock In \emph{ICCV}, pages 5127--5137, 2019{\natexlab{b}}.

\bibitem[Zhang et~al.(2024{\natexlab{b}})Zhang, Qiu, Zhou, and An]{Zhang_2024_ACCV}
Qianqian Zhang, Linwei Qiu, Li Zhou, and Junshe An.
\newblock Esm-yolo: Enhanced small target detection based on visible and infrared multi-modal fusion.
\newblock In \emph{ACCV}, pages 1454--1469, 2024{\natexlab{b}}.

\bibitem[Zhao et~al.(2024{\natexlab{a}})Zhao, Cai, Dong, and Hu]{10655245}
Chen Zhao, Weiling Cai, Chenyu Dong, and Chengwei Hu.
\newblock Wavelet-based fourier information interaction with frequency diffusion adjustment for underwater image restoration.
\newblock In \emph{CVPR}, 2024{\natexlab{a}}.

\bibitem[Zhao et~al.(2024{\natexlab{b}})Zhao, Ye, and Du]{s24134098}
Pujie Zhao, Xia Ye, and Ziang Du.
\newblock Object detection in multispectral remote. sens. images based on cross-modal cross-attention.
\newblock \emph{Sensors}, 2024{\natexlab{b}}.

\bibitem[Zhao et~al.(2024{\natexlab{c}})Zhao, Yuan, Jiang, Wang, and Wei]{zhao2024removalselectioncoarsetofinefusion}
Tianyi Zhao, Maoxun Yuan, Feng Jiang, Nan Wang, and Xingxing Wei.
\newblock Removal then selection: A coarse-to-fine fusion perspective for rgb-infrared object detection.
\newblock \emph{arXiv:2401.10731}, 2024{\natexlab{c}}.

\bibitem[Zhao et~al.(2020)Zhao, Xu, Zhang, Liu, Zhang, and Li]{Zhao_2020}
Zixiang Zhao, Shuang Xu, Chunxia Zhang, Junmin Liu, Jiangshe Zhang, and Pengfei Li.
\newblock Didfuse: Deep image decomposition for infrared and visible image fusion.
\newblock In \emph{IJCAI}, 2020.

\bibitem[Zhao et~al.(2022)Zhao, Xu, Zhang, Liang, Zhang, and Liu]{9416456}
Zixiang Zhao, Shuang Xu, Jiangshe Zhang, Chengyang Liang, Chunxia Zhang, and Junmin Liu.
\newblock Efficient and model-based infrared and visible image fusion via algorithm unrolling.
\newblock \emph{TCSVT}, 2022.

\bibitem[Zhao et~al.(2023)Zhao, Bai, Zhang, Zhang, Xu, Lin, Timofte, and Van~Gool]{Zhao_2023_CVPR}
Zixiang Zhao, Haowen Bai, Jiangshe Zhang, Yulun Zhang, Shuang Xu, Zudi Lin, Radu Timofte, and Luc Van~Gool.
\newblock Cddfuse: Correlation-driven dual-branch feature decomposition for multi-modality image fusion.
\newblock In \emph{CVPR}, 2023.

\bibitem[Zhao et~al.(2024{\natexlab{d}})Zhao, Bai, Zhang, Zhang, Zhang, Xu, Chen, Timofte, and Van~Gool]{Zhao_2024_CVPR}
Zixiang Zhao, Haowen Bai, Jiangshe Zhang, Yulun Zhang, Kai Zhang, Shuang Xu, Dongdong Chen, Radu Timofte, and Luc Van~Gool.
\newblock Equivariant multi-modality image fusion.
\newblock In \emph{CVPR}, 2024{\natexlab{d}}.

\bibitem[Zhou et~al.(2024)Zhou, Li, Qiao, Xie, Wang, Ruan, Mei, and Yang]{zhou2024dmmdisparityguidedmultispectralmamba}
Minghang Zhou, Tianyu Li, Chaofan Qiao, Dongyu Xie, Guoqing Wang, Ningjuan Ruan, Lin Mei, and Yang Yang.
\newblock Dmm: Disparity-guided multispectral mamba for oriented object detection in remote. sens.
\newblock \emph{arXiv:2407.08132}, 2024.

\bibitem[Zhu et~al.(2025)Zhu, Zhang, Li, Wang, and Ma]{10770223}
Jiahe Zhu, Huan Zhang, Simin Li, Shengjin Wang, and Hongbing Ma.
\newblock Cross teaching-enhanced multispectral remote sensing object detection with transformer.
\newblock \emph{J-STARS}, 18:\penalty0 2401--2413, 2025.

\bibitem[Zhu et~al.(2023)Zhu, Sun, Wang, and Huang]{10105844}
Yaohui Zhu, Xiaoyu Sun, Miao Wang, and Hua Huang.
\newblock Multi-modal feature pyramid transformer for rgb-infrared object detection.
\newblock \emph{IEEE Trans. Intell. Transport. Syst.}, 24\penalty0 (9):\penalty0 9984--9995, 2023.

\bibitem[Zou et~al.(2021)Zou, Jiang, Zhang, Chen, Lu, and Wu]{Zou_2021_ICCV}
Wenbin Zou, Mingchao Jiang, Yunchen Zhang, Liang Chen, Zhiyong Lu, and Yi Wu.
\newblock Sdwnet: A straight dilated network with wavelet transformation for image deblurring.
\newblock In \emph{ICCVW}, 2021.

\end{thebibliography}
}

\renewcommand\thefigure{\Alph{figure}} 
\setcounter{figure}{0}
\renewcommand\thetable{\Alph{table}} 
\setcounter{table}{0}
\renewcommand{\thesection}{\Alph{section}}
\setcounter{section}{0}

\setcounter{page}{1}
\maketitlesupplementary
In this supplementary file, we introduce the details of the datasets and evaluation metrics in Sec.~\ref{sec:data}, and the implementation in Sec.~\ref{sec: implement details}. We also provide experimental results on the dataset VEDAI and KAIST in Sec.~\ref{sec:experiment}. Additionally, we provide more visualization results and ablation studies in Sec.~\ref{sec: visualization results} and Sec.~\ref{Ablation study2}, respectively.

\begin{table*}
  \centering
  \scalebox{0.85}{
  \begin{tabular}{cccccc}
    \toprule
    Methods&Backbone&$mAP_{50}$ & $mAP$ & Parameters & Inference time (ms)\\
    \midrule
     Ours&ResNet50&\textcolor{blue}{87.9}&\textcolor{blue}{59.4}&193.2M&53.2\\
     \midrule
     (TNNLS'23) LRAF-Net~\cite{10144688}&YOLOv5&85.9&59.1&-&-\\
     (PR'24) ICAFusion~\cite{DBLP:journals/pr/ShenCLZFY24}&YOLOv5&84.8&56.6&164.3M&\textcolor{blue}{50.2}\\
     (GRSL'24) SSE+FFT~\cite{xu2024enhanced}&YOLOv5&86.5&56.9&-&-\\
      (ACCV'24) ESM-YOLO~\cite{Zhang_2024_ACCV}&YOLOv5&82.4&-&80.2M&-\\
      (JSTAEORS'25) DCCCNet~\cite{10797700}&YOLOv5&82.0&49.9&-&-\\
      (GRSL'24) CrossYOLO~\cite{10342625}&YOLOv7&79.8&-&-&-\\
      Ours&YOLOv5&\textcolor{green}{88.2}&\textcolor{green}{59.6}&\textcolor{red}{45.6M}&\textcolor{green}{44.1}\\
      \midrule
      YOLOv8l-IR&YOLOv8&73.6&52.2&43.7M&22.0\\
      YOLOv8l-RGB&YOLOv8&62.9&46.5&43.7M&22.0\\
      \midrule
      (JSTAEORS'25) MDA~\cite{10770223}&DETR&82.2&-&\textcolor{blue}{72.4M}&-\\
     (ASC'24) DSM-AVD~\cite{ZHANG2024111971}&YOLOv8&79.3&50.3&-&-\\
     Ours&YOLOv8&\textcolor{red}{88.4}&\textcolor{red}{59.8}&\textcolor{green}{69.1M}&\textcolor{red}{40.0}\\
    \bottomrule
  \end{tabular}}
  \caption{Comparison results with SOTA methods on VEDAI dataset. The best, second and third results are highlighted in \textcolor{red}{red}, \textcolor{green}{green} and \textcolor{blue}{blue}, respectively.}
  \label{tab:VEIDA1fold}
  \vspace{-10pt}
\end{table*}
\begin{table}
  \centering
  \small
  \scalebox{1.0}{
  \begin{tabular}{cc|cc}
    \toprule
    Methods&Backbone&$mAP_{50}$ & $mAP$\\
    \midrule
    (CVPRW'19) MMTOD~\cite{devaguptapu2019borrow}&ResNet50&70.7&31.3\\
    (WACV'21) GAFF~\cite{Zhang_2021_WACV}&ResNet50&67.1&24.4\\
    (TCSVT'22) CMDet~\cite{9759286}&ResNet50&68.4&28.3\\
    Ours & ResNet50&\textcolor{blue}{75.0}&\textcolor{blue}{34.2}\\
    \midrule
    (2021) CFT~\cite{qingyun2021cross}&YOLOv5&71.2&29.3\\
    (RS’22) RISNet~\cite{wang2022improving}&YOLOv5&72.7&33.1\\
    (PR'24) ICAFusion~\cite{DBLP:journals/pr/ShenCLZFY24}&YOLOv5&60.3&-\\
    (CVPRW'24) DaFF~\cite{10677953}&YOLOv5&61.9&-\\
    Ours & YOLOv5&\textcolor{green}{75.4}&\textcolor{green}{34.4}\\
    \midrule
    YOLOv8l-IR~\cite{10533619}&YOLOv8&56.8&22.4\\
    YOLOv8l-RGB~\cite{10533619}&YOLOv8&55.3&21.6\\
    \midrule
    (Sensors'23) Dual-YOLO~\cite{s23062934}&YOLOv7&73.2&-\\
    (Sensors'24) IV-YOLO~\cite{s24196181}&YOLOv8&75.4&-\\
    Ours & YOLOv8&\textcolor{red}{75.8}&\textcolor{red}{34.8}\\
    \bottomrule
  \end{tabular}}
  \caption{Comparison results with SOTA methods on KAIST dataset. The best results are highlighted in \textcolor{red}{red}. The second and third best results are
highlighted in \textcolor{green}{green} and \textcolor{blue}{blue}, respectively.}
\label{tab:KAIST}
\end{table}

 \begin{figure*}[tbp]
     \centering \includegraphics[width=1.0\linewidth]{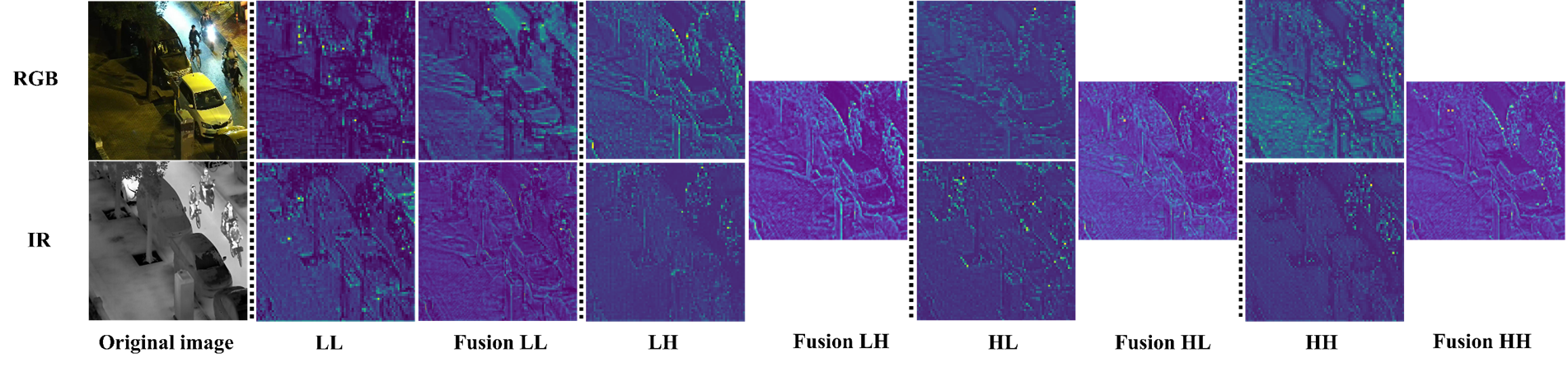}
     \caption{
     The illustration shows how the original features are filtered in two directions (horizontal and vertical) using DWT, resulting in four sub-bands: LL (low-low), LH (low-high), HL (high-low), and HH (high-high). "Fusion" refers to the results obtained after combining these sub-bands through our method.
     } 
     \label{fig:placeholder6} 
 \end{figure*}
 \begin{figure}[tbp]
    \centering \includegraphics[width=0.5\textwidth,height=0.25\textheight]{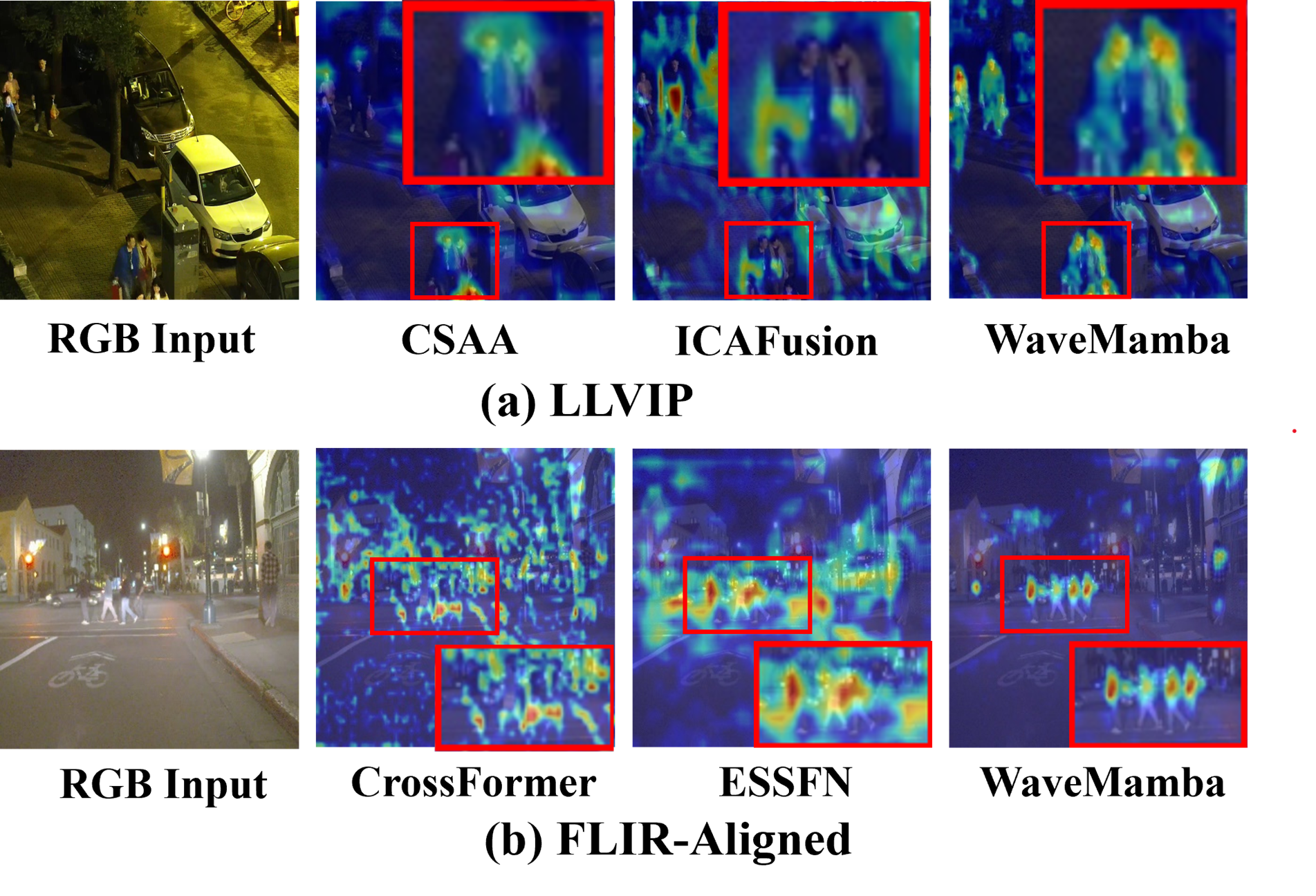}
    \caption{Heatmap visualization of several cross-modality object detection methods on LLVIP and FLIR-Aligned.} 
    \label{fig:placeholder4} 
\end{figure}
\begin{table}
  \centering
  \scalebox{0.9}{
  \begin{tabular}{cccc}
    \toprule
    Methods&$mAP_{50}$ & $mAP$ & Parameters\\
    \midrule
    \{$P_1$,$P_2$,$P_3$\}& 91.1 &62.8 &23.1M\\
    \{$P_2$,$P_3$,$P_4$\}& 91.8 &63.5 &58.3M\\
    \{$P_3$,$P_4$,$P_5$\}&91.5&62.9&84.6M\\
    \{$P_2$,$P_4$,$P_5$\}& 91.3 &62.5 &78.3M\\
    \{$P_2$,$P_3$,$P_5$\} (ours)&92.1&64.4&69.1M\\
    \bottomrule
  \end{tabular}}
  \caption{Effects of positions of WaveMamba Fusion Blocks on $M^{3}$FD dataset.}
  \label{tab:example9}
\end{table}
\begin{table}
  \centering
  \scalebox{0.9}{
  \begin{tabular}{cccc}
    \toprule
    Methods&$mAP_{50}$ & $mAP$ & Parameters\\
    \midrule
    (One Blocks)& 91.3&63.8 &59.3M\\
    (Two Blocks)&91.6&64.0&61.6M\\
    (Three Blocks) (ours)&92.1&64.4&69.1M\\
    \bottomrule
  \end{tabular}}
  \caption{Effects of different number of WaveMamba Fusion Blocks on $M^{3}$FD dataset.}
  \label{tab:example13}
\end{table}
\begin{table}
  \centering
  \scalebox{0.9}{
  \begin{tabular}{cccc}
    \toprule
    Methods&$mAP_{50}$ & $mAP$ & Parameters\\
    \midrule
    (MEYR)& 91.7 &64.0 &69.1M\\
    (SYM3)& 91.8 &64.1 &69.1M\\
    (COIF3)& 91.7&64.1 &69.1M\\
    (DB3)&91.8&64.2&69.1M\\
    (HAAR) (ours)&92.1&64.4&69.1M\\
    \bottomrule
  \end{tabular}}
  \caption{Effects of different wavelet bases in WaveMamba Fusion Blocks on $M^{3}$FD dataset.}
  \label{tab:example11}
\end{table}
\begin{figure*}[tbp]
    \centering \includegraphics[width=1.0\linewidth]{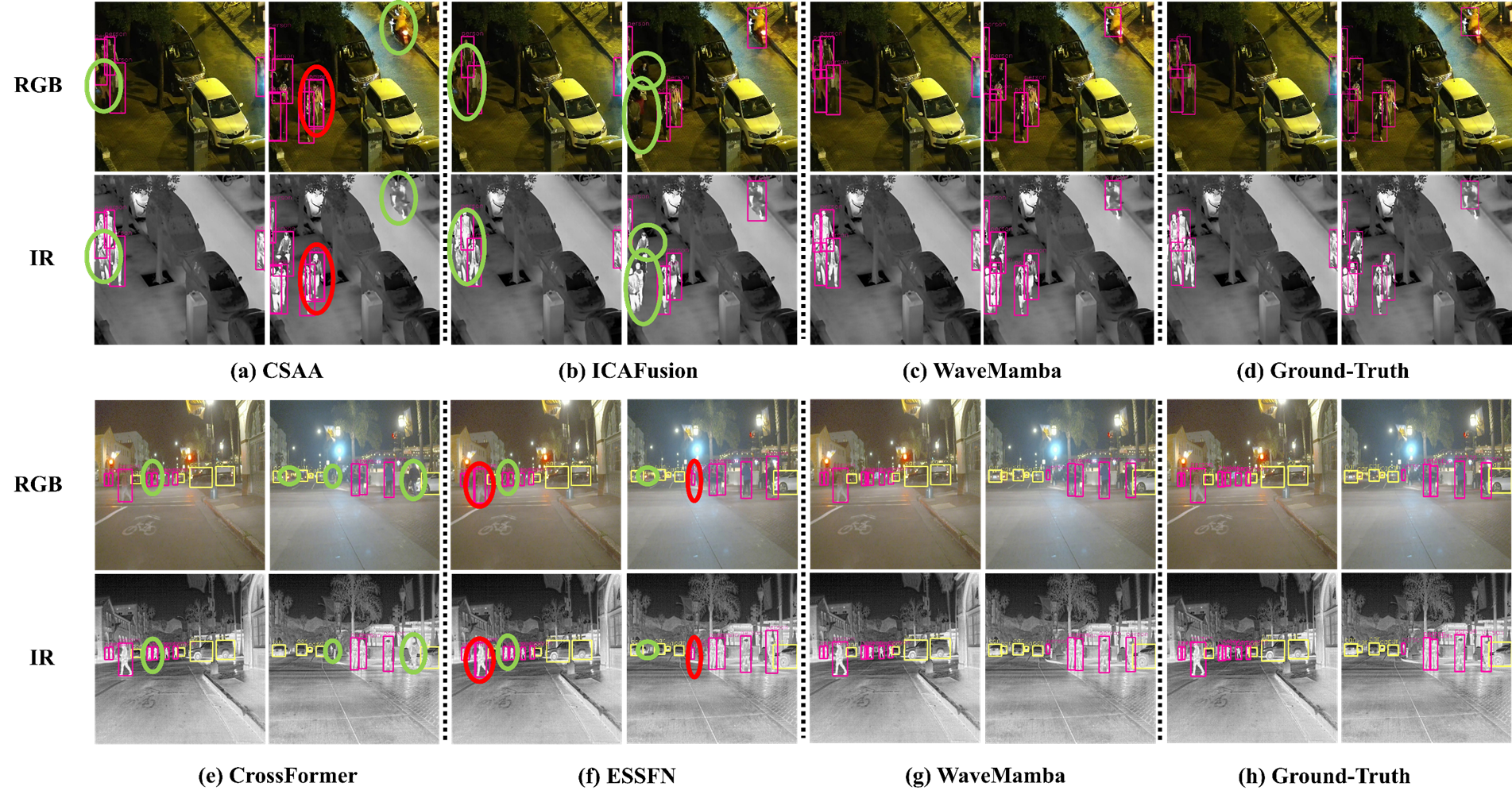}
    \caption{Detection results' visualization of several cross-modality object detection methods on LLVIP and FLIR-Aligned. Wherein, (a)-(d) present the results of LLVIP dataset, and (e)-(h) present the results of FLIR-Aligned dataset. The targets encircled by red ellipses are false positives, while those encircled by green ellipses are missed detections. Please zoom in for more details.} 
    \label{fig:placeholder5} 
\end{figure*}

\section{Datasets and Evaluation Metrics}
\label{sec:data}
\textbf{Experimental Datasets.} We evaluate our methods on four common-used visible-infrared object detection benchmark datasets : $M^{3}$FD Dataset~\cite{9879642}, DroneVehicle Dataset~\cite{9759286}, LLVIP Dataset~\cite{9607632}, FLIR Dataset~\cite{reference1} and two additional datasets: VEDAI Dataset~\cite{razakarivony2016vehicle} and KAIST Dataset~\cite{hwang2015multispectral}.\\
\indent $M^{3}$FD dataset is a benchmark dataset for multi-class RGB-IR detection which collects $4,200$ pairs of aligned images from various scenes. These pictures captured under low-light conditions or in adverse weather, pose a significant challenge to the detection performance of the model. This dataset encompasses six categories of objects that frequently appear in autonomous driving or road surveillance scenarios. Since it does not provide an official criterion for dividing the training set and validation set, we adopt the division standard from \cite{Liang2023ExplicitAF} as $3,360$ pairs for training and $840$ pairs for testing.\\
\indent DroneVehicle dataset is a large-scale RGB-IR vehicle detection dataset, consisting of $28,439$ pairs of images and $953,087$ annotations for five categories: \textit{car}, \textit{truck}, \textit{bus}, \textit{van}, and \textit{freight-car}. The images are collected by drones under varying lighting conditions, angles, and altitudes. The rich perspective variations combined with a large number of dense annotations make it difficult for models to achieve high detection performance on this dataset. Following the official method for dataset partitioning, we use $17,990$ image pairs for training, $1,469$ pairs for validation, and $8,980$ pairs for testing. We report the testing part's results.\\
\indent LLVIP dataset is an aligned low-light RGB-IR dataset which is specially collected for pedestrian detection with $15,488$ image pairs. According to the official standard, we use $12,025$ image pairs for training and $3,463$ pairs for testing.\\
\indent FLIR dataset is a relatively difficult RGB-IR detection dataset with five categories: \textit{people}, \textit{car}, \textit{bike}, \textit{dog} and \textit{other cars}. Due to the low-quality annotations and a large number of unaligned image pairs in the original dataset, following by~\cite{zhang2020multispectral}, we adopt the FLIR-Aligned dataset which includes $4,129$ pairs of images for training and $1,013$ pairs for testing. We remove the `\textit{dog}' category from the dataset due to its few number of instances.\\
\indent VEDAI dataset is a remote sensing detection dataset consisting of $1,210$ pairs of RGB-IR images captured from a drone at high altitudes with nine different types of objects such as \textit{car}, \textit{truck} and \textit{pickup}. Since the objects are predominantly small targets, it presents a significant challenge to the detection performance of the model. Since the dataset does not have an official split, we follow the common training and testing set partitioning methodology with $1,089$ pairs for training and $121$ pairs for testing.\\
\indent KAIST dataset is a public low-light multi-spectral pedestrian detection dataset. Due to problems in the original dataset annotations, we utilize the improved training~\cite{zhang2019weakly} and testing annotations~\cite{liu2016multispectral} that are widely adopted by researchers. Following the most commonly used data partitioning method provided by~\cite{zhang2019weakly}, we use $8,963$ image pairs for training and $2,252$ pairs for testing.\\
\textbf{Evaluation metrics.} Since our task is object detection, we choose the most widely used metrics $mAP_{50}$ and $mAP$ to evaluate the performance of models on six datasets. The $mAP_{50}$ metric represents the mean AP under IoU $0.50$ and the $mAP$ metric represents the mean AP under IoU ranges from $0.50$ to $0.95$ with a stride of $0.05$~\cite{zhao2024removalselectioncoarsetofinefusion}. For the multi-class datasets $M^{3}$FD and DroneVehicle, we also provide the $AP_{50}$ results for each category. Due to the high difficulty of FLIR-Aligned, we additionally report the results of precision, recall, and F1 score. All the evaluation metrics indicate better model detection performance when their values are higher. We also present the average inference time of our method, evaluated on an A800 GPU over $15$ runs using input image pairs of size $640\times640$. Additionally, we provide the parameter specifications of our model.
\section{Implementation Details}
\label{sec: implement details}
All experiments on our six datasets are conducted on a single A800 GPU, with a batch size of $16$ during training and $32$ during testing. The input image pairs' size for both testing and training are $640\times640$ and the training epoch is set to $250$ for all four datasets with an initial learning rate of $0.01$. We utilize the SGD optimizer with a momentum of $0.937$ and a weight decay of $0.0005$. The loss function, other hyper-parameters, and data augmentation parameters all adopt the default settings of the original YOLOv8~\cite{10533619}.
\section{More experiments on VEDAI and KAIST}
\label{sec:experiment}
\noindent\textbf{VEIDA Dataset.} The results on VEDAI are summarized on Table~\ref{tab:VEIDA1fold}. Our method achieves top-three rankings on both $mAP_{50}$ and $mAP$, surpassing the fourth-place method by $1.9$\% and $0.7$\% and achieving $88.4$\% and $59.8$\%, respectively. Moreover, our method, based on the YOLOv5 and YOLOv8 backbones, has the smallest number of parameters, requiring only $44.1$ ms and $40.0$ ms to process a pair of images, respectively.\\
\noindent\textbf{KAIST Dataset.} 
Table~\ref{tab:KAIST} shows the results of our method with other SOTA methods on KAIST. Our method achieves the top three results using three different backbones on both $mAP_{50}$ and $mAP$, surpassing the fourth method by $0.4$\% and $1.7$\%, respectively. Notably, although the feature extraction capability of the YOLOv5 backbone is inferior to that of YOLOv7, our method using YOLOv5 still outperforms Dual-YOLO, which utilize YOLOv7. The results demonstrate the superior performance of our WMFBs.
\section{Visualization}
\label{sec: visualization results}
\subsection{Frequency-domain graph}
We visualize the wavelet transform outputs of RGB and IR features derived from the second stage of YOLOv8 backbones trained on each respective modality from a pair of RGB and IR images, as well as the high- and low-frequency components after fusion. 
As shown in Fig.~\ref{fig:placeholder6}, the low-frequency components of IR more effectively convey shape information compared to RGB, while the high-frequency components of RGB more distinctly emphasize local object contours and details than IR.
After fusion, the low-frequency components of both RGB and IR become clearer, and the local object features in the high-frequency components after fusion are significantly enhanced. 

\subsection{Heatmaps}
\label{heatmaps}
Based on Grad-CAM~\cite{8237336}, we visualize the heatmaps of our model's first inverse wavelet transform layer (for the enhanced high- and low-frequency features obtained at the third fusion block) and compare them with other state-of-the-art methods~\cite{10209020,DBLP:journals/pr/ShenCLZFY24,LEE2024144,10653751} on LLVIP
and FLIR-Aligned. As shown in Fig.~\ref{fig:placeholder4}, our model focuses more intently on detecting targets without being excessively distracted by background noise by utilizing wavelet transform and feature fusion to prioritize the targets. In contrast, other methods either exhibit excessive attention dispersion in background areas or cover the detection targets with a wide range of attention, which easily leads to missed detections of targets within the region.
\subsection{Detection Results}
We visualize the detection results and compare them with several SOTA methods~\cite{10209020,DBLP:journals/pr/ShenCLZFY24,LEE2024144,10653751}. As presented in Fig.~\ref{fig:placeholder5}, under low-light or heavily occluded conditions, our method has reduced the number of missed and false detected targets compared to other methods. It successfully detects more difficult targets and achieves the best detection performance.
\section{Ablation Study}
\label{Ablation study2}
\subsection{Details of the Improved Head}
To demonstrate the effectiveness of our improved YOLOv8 head, we also combine our fusion modules with the original YOLOv8 head for comparison. To seamlessly integrate our method into the original YOLOv8 head, we sum the fused low-frequency components obtained after each fusion layer, concatenate them with the fused high-frequency components for inverse wavelet transformation, and then feed the result into the original YOLOv8 head.
\subsection{More Ablation Experiments}
\noindent\textbf{Effects of WMFBs' positions.} Like previous works~\cite{LEE2024144,DBLP:journals/corr/abs-2111-00273}, we also employ three feature fusion blocks.
Table~\ref{tab:example9} shows the effect of different position combinations of WMFBs on performance. $P_i$ represents that the WMFB is placed at $i^{th}$ stage. 
Based on the experimental results from ``\{ $P_1$, $P_2$, $P_3$\}''  and~\cite{LIU2017191}, the first layer features are not suitable for fusion.
Without $P_1$, there are four remaining possible combinations of layers.
When comparing ``\{ $P_2$, $P_3$, $P_5$\}'' with others, the importance of using features from the second, third, and fifth layers is verified by the significant accuracy boost.\\
\noindent\textbf{Effects of the number of WMFBs.} Since YOLOv8 head inherently contains three detection modules, we conduct an ablation study on the effects of the number of WMFBS by changing one or two WMFBs to simple AVG method and the results are shown in Table~\ref{tab:example13}. Reducing the number of WMFBs from three to one and two results in a decrease of $0.8$\% and $0.5$\% in $mAP_{50}$ and a decrease of $0.6$\% and $0.4$\% in $mAP$, respectively.\\
\noindent\textbf{Effects of different Wavelet Bases.} 
We show the influence of different wavelet bases in Table~\ref{tab:example11}.
The performance difference in $mAP_{50}$ and $mAP$ is less than $0.4\%$ and the Haar wavelet base achieves the best performance. 
This result indicates that our method is not sensitive to the selection of wavelet bases.

\end{document}